\documentclass{article}

\usepackage[preprint]{neurips_2025}
\usepackage{comment}
\usepackage{graphicx}
\usepackage{multirow}

\usepackage[utf8]{inputenc} 
\usepackage[T1]{fontenc}    
\usepackage{hyperref}       
\usepackage{url}            
\usepackage{booktabs}       
\usepackage{amsfonts}       
\usepackage{nicefrac}       
\usepackage{microtype}      
\usepackage{xcolor}         
\usepackage{amsmath}

\usepackage[backend=biber,style=numeric]{biblatex} 
\title{Mollifier Layers: Enabling Efficient High-Order Derivatives in Inverse PDE Learning}

\author{%
  Ananyae Kumar Bhartari\thanks{Equal Contribution} \\
  University of Pennsylvania\\
  \texttt{ananyae@seas.upenn.edu} \\
  \And
  Vinayak Vinayak\footnotemark[1] \\
  University of Pennsylvania\\
  \texttt{vinayakv@seas.upenn.edu} \\
  \AND
  Vivek B Shenoy \\
  University of Pennsylvania\\
  \texttt{vshenoy@seas.upenn.edu} \\
}

\begin{document}

\maketitle

\begin{abstract}
  Parameter estimation in inverse problems involving partial differential equations (PDEs) underpins modeling across scientific disciplines, especially when parameters vary in space or time. Physics-informed Machine Learning (PhiML) integrates PDE constraints into deep learning, but prevailing approaches depend on recursive automatic differentiation (autodiff), which produces inaccurate high-order derivatives, inflates memory usage, and underperforms in noisy settings. We propose Mollifier Layers, a lightweight, architecture-agnostic module that replaces autodiff with convolutional operations using analytically defined mollifiers. This reframing of derivative computation as smoothing integration enables efficient, noise-robust estimation of high-order derivatives directly from network outputs. Mollifier Layers attach at the output layer and require no architectural modifications. We compare them with three distinct architectures and benchmark performance across first-, second-, and fourth-order PDEs—including Langevin dynamics, heat diffusion, and reaction-diffusion systems—observing significant improvements in memory efficiency, training time and accuracy for parameter recovery across tasks. To demonstrate practical relevance, we apply Mollifier Layers to infer spatially varying epigenetic reaction rates from super-resolution chromatin imaging data—a real-world inverse problem with biomedical significance. Our results establish Mollifier Layers as an efficient and scalable tool for physics-constrained learning.
  
\end{abstract}

\section{Introduction}

Parameter estimation through inverse problems constrained by PDEs arise across science and engineering, where parameters like diffusivity or conductance must be inferred from limited, noisy measurements\cite{kaipio2006statistical,tarantola2005inverse,Herrera_param_hydro,Zhang_ocean}. Traditional inference methods are computationally intensive, as they require repeatedly solving forward PDEs. These challenges grow worse in high-resolution or real-time settings, especially when governing parameters vary at fine scales or data is sparse\cite{Xun_param_est}.

Deep learning has emerged as a powerful framework for such problems, with physics-informed machine learning (PhiML) approaches\cite{Toscano_review}—like PINNs\cite{Raissi}, DeepFNO\cite{Li_FNO}, and DeepONet\cite{Lu_FNO}—embedding PDE constraints into model training. These methods enable learning from limited observations, but rely heavily on recursive automatic differentiation (autodiff) to compute spatial and temporal derivatives\cite{baydin2018automatic}. Autodiff based PhiML techniques are memory-intensive and unstable for higher-order derivatives\cite{Wang_piratenet,Fucks_limitations,fourment2023automatic,margossian2019review}, making such models brittle under noise and costly for deep architectures. 

To overcome these limitations, we propose \textit{mollifier layers}\cite{Friedrichs_weakstrong,murio1998discrete}: lightweight, architecture-agnostic modules that replace autodiff with convolution-based derivative estimation using smooth, compactly supported kernels. These layers estimate derivatives via convolution with analytic derivatives of smooth mollifier kernels, transferring differentiation off the network and implicitly regularizing the computation for stability and noise robustness.

Integrated into and benchmarked against three PhiML architectures—PINNs\cite{Raissi}, PirateNet\cite{Wang_piratenet}, and PINNsFormer\cite{zhao2023pinnsformer}—mollifier layers offer a scalable, robust alternative to autodiff for physics-informed learning. They achieve significant improvements in parameter recovery, derivative stability, and computational efficiency across diverse PDEs, including a fourth-order reaction-diffusion model\cite{Heo} where they accurately infer the governing parameter while reducing the memory usage and time complexity by over an order of magnitude. Applied to super-resolution microscopy\cite{schermelleh2019super} of the human cell nuclei, they enable spatially resolved inference of critical biophysical parameters from noisy data, demonstrating practical utility in real-world scientific modeling.

While our benchmarks target inverse PDEs, the underlying principle—offloading high-order derivative computation to analytic mollifier convolutions—can generalize to forward models\cite{karniadakis2021physics} (shown briefly in B.5), operator learning\cite{azizzadenesheli2024neural}, or neural ODE systems\cite{ruiz2023neural,chen2018neural} where gradient accuracy is crucial.

\begin{figure}
  \centering
  \includegraphics[width=\linewidth]{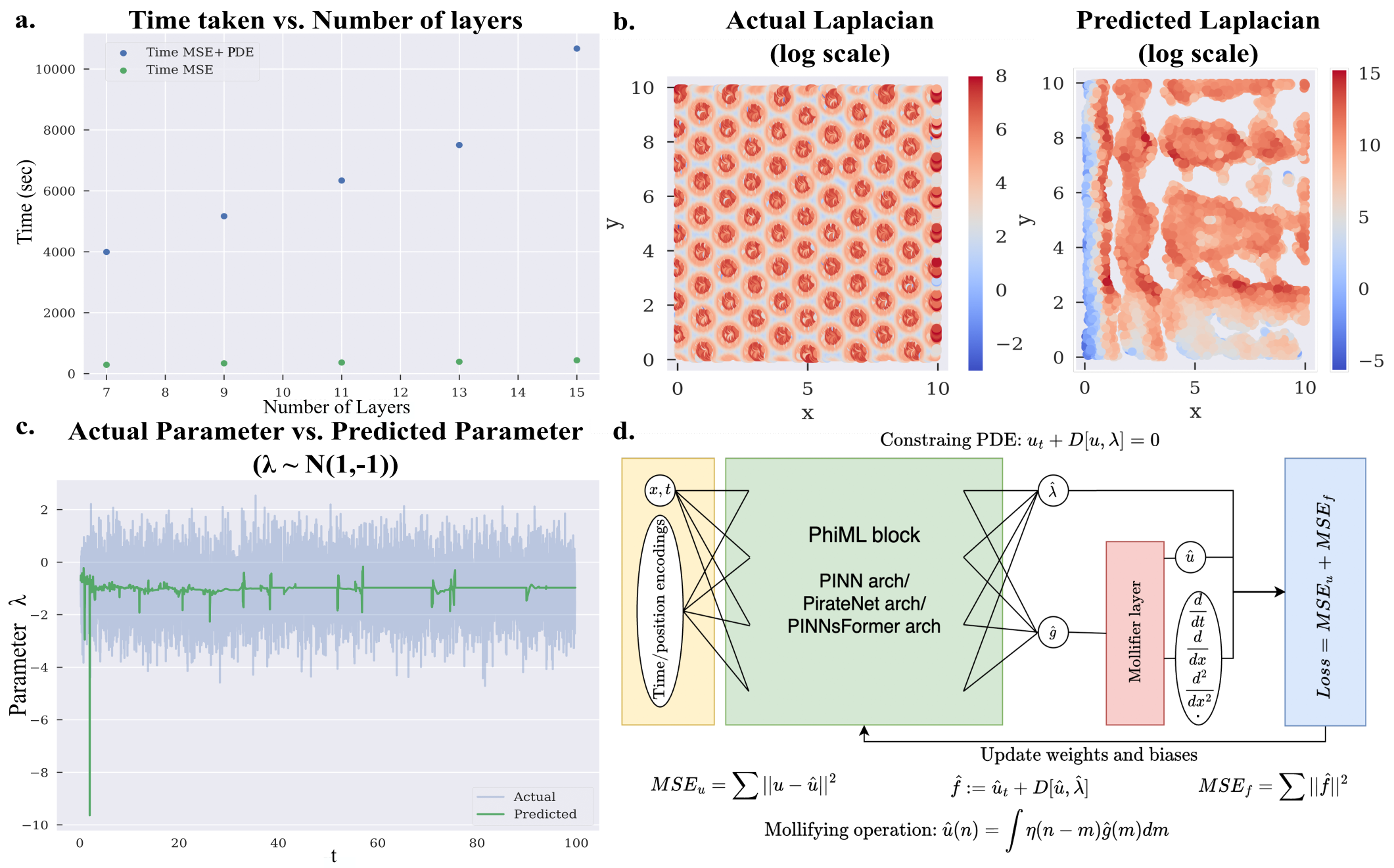}
  \caption{Limitations of autodiff and overview of PhiML+Mollifier architecture. \textbf{(a)} Training time comparison for PINNs with and without PDE residual loss. \textbf{(b)} PINN-predicted vs. actual Laplacian in a reaction-diffusion system. \textbf{(c)} PINN-predicted vs. actual forcing term in a Langevin system. \textbf{(d)} PhiML+Mollifier Layer architecture replacing autodiff with mollifier-based convolution.}
\end{figure}

\section{Motivation}
As shown in prior work\cite{Wang_piratenet,Fucks_limitations,fourment2023automatic,margossian2019review}, recursive autodiff—despite its central role in PhiML—poses key bottlenecks for learning PDEs, particularly in the accurate and efficient estimation of higher-order derivatives. To expose and quantify these, we benchmarked an upgraded PINN (details in A.1) on spatio-temporally varying parameter estimation. This analysis revealed three core issues—compute cost, derivative inaccuracy, and memory blow-up—that directly motivated our mollifier layer design.

\textbf{Motivation \#1: Rising cost of high-order derivatives with model depth.} Evaluating PDE residuals in PhiML requires multiple backward passes for each order of differentiation, causing compute time to grow superlinearly with network depth\cite{Cho_PINN,Cardona_MLST}. This becomes especially burdensome in complex systems that demand deeper networks. For instance, in the reaction-diffusion benchmark (Sec. 4.3), training PINNs with 7 to 15 layers using the full loss (data + PDE) took significantly longer than using the data loss alone—an overhead that widens with depth (Fig. 1a). Such costs slow down training and escalate energy and hardware requirements, forming a major scalability barrier.

\textbf{Motivation \#2: Inaccuracy in higher-order derivatives for high-frequency features.} Recursive autodiff is susceptible to rounding errors from repeated gradient operations. This instability becomes particularly problematic when modeling spatio-temporally varying or noisy parameters. For example, a PINN trained to recover the gradient in a reaction-diffusion system achieved only 0.21 Laplacian correlation with ground truth (Fig. 1b), indicating poor quantitative accuracy.

These issues are exacerbated with high-frequency or noisy signals, where the inherent smoothness of neural networks hinders the capture of sharp or discontinuous features\cite{Fucks_limitations}. In a Langevin system with Gaussian white noise (Sec. 4.1.1), the PINN failed to capture variance in the inferred parameter (Fig. 1c). Together, these results underscore the fragility of autodiff in noisy, high-order settings, motivating the need for more stable and noise-resilient derivative estimation techniques.

\begin{table}
  \caption{Peak Memory usage by PINNs}
  \label{sample-table}
  \centering
\begin{tabular}{lcccr}
\toprule
Equation & MSE loss only & MSE loss + PDE loss \\
\midrule
Langevin (first order 1D)  & 0.17 GB & 0.21 GB\\
Heat (second order 2D) & 0.8 GB & 1.2 GB\\
Reaction-Diffusion (fourth order 2D)     & 0.4 GB & 2.7 GB \\
\bottomrule
\end{tabular}
\end{table}

\textbf{Motivation \#3: High memory overhead from recursive autodiff.} Computing higher-order derivatives through recursive autodiff requires storing intermediate gradients across layers, leading to memory usage that scales with both depth and derivative order. In experiments with parameter estimation for Langevin, heat, and reaction-diffusion equations using PINN(Sec. 4), adding the PDE error significantly increased GPU memory usage (hardware specs in A.6) compared to training with data-only losses (Table 1). This memory blow-up limits scalability and highlights the need for more memory-efficient alternatives.

\section{Methods}

\subsection{Formulating parameter inference using PhiML}
To ground our innovations, we begin with a brief overview of parameter inference for PDEs using a general PhiML framework. Without loss of generality, consider a partial differential equation:
\begin{equation}
    u_t+D[u,\lambda]=0,x\in \Omega,t\in T
\end{equation}
where $u(t,x)$ is the solution of the PDE, $D[.;\lambda]$ is a non-linear/linear ordinary or partial differential operator parameterized by $\lambda(t,x)$ and the domain $\Omega$ is a subset in $R^n$. Subscripts denote differentiation in time $t$ or space $x$. Given this PDE, the aim is to infer $\lambda(t,x)$ given a sparsely sampled set of $u(t,x)$. In PhiML, the solution  $u(t,x)$ is approximated by a deep neural network $u_\theta (t,x)$ (or $\hat{u}$),  where $\theta$ represents the set of all trainable parameters of the network. With the autodiff formulation\cite{Paszke_autodiff}, required gradients with respect to input variables $(t,x)$ or network parameters $\theta$ can be computed. Additionally, $\hat{\lambda}$, the estimated parameter, is also an output of the neural network. The actual parameter $\lambda(t,x)$ is spatially or temporally varying, which can be due to a pattered modulation in these parameters or fluctuations due to noise. We assume that $\lambda(t,x)$ is unknown across $\Omega$, making parameter inference challenging but broadly applicable. 

Combining the estimates $(\hat{u}, \hat{\lambda})$ with the derivatives of $\hat{u}$ computed via autodiff, the PDE residual is defined as $\hat{f} = \hat{u}_t + D[\hat{u},\hat{\lambda}]$, evaluated at all points in $\Omega$. With these tools, the optimization problem is to minimize the total mean squared error $(MSE_{total})$, which is defined as $MSE_{total}=MSE_{u}+MSE_{f},$ where,
\begin{equation}
    \text{Data loss: }MSE_{u}=\frac{1}{N_u}  \sum_{i=N_u}|\hat{u}(t^i,x^i )-u(t^i,x^i)|^2 
\end{equation}
 and
\begin{equation}
    \text{PDE residual loss: }MSE_{f}=\frac{1}{N_f} \sum_{j=N_f}|\hat{f}(t^j,x^j,\lambda^j)|^2.
\end{equation}

Here, $N_u$ are the training data points available for the actual solution $u(t,x)$ and $N_f$ all all the points in the discretized domain for which the solution is being sought. For the initial testing of the models and characterizing their behavior, we use $N_u=N_f$ and later validate the models on $N_u=0.1N_f$, i.e., with 10\% sampling. The data loss is calculated on $N_u$, while the PDE loss on $N_f$.

\subsection{Modified parameter inference with Mollifier Layers}

\underline{\textit{Intuition behind Mollifier Layers:}} Recognizing the limitations of recursive autodiff, we draw from the weak-form formulation in finite element methods\cite{hughes2003finite}, where derivatives are inferred via integration against smooth test functions rather than evaluated pointwise. This principle suggests an alternative to gradient chaining: recovering derivatives through structured integration applied directly to the network output. A similar idea underlies Savitzky–Golay filters\cite{savitzky1964smoothing}, which compute stable numerical derivatives by fitting local polynomials within a sliding window—smoothing and differentiating simultaneously to mitigate the effects of noise and discretization.

Mollifier layers formalize this intuition within neural architectures: rather than relying on recursive autodiff, they compute derivatives via convolution with analytic derivatives of smooth, compactly supported kernels. These mollifiers serve as localized test functions, converting differentiation into a stable smoothing operation that is mathematically grounded and memory-efficient. Compared to alternatives like Gaussian filters or splines, mollifiers offer two key advantages: (1) compact support for localized and efficient computation, and (2) closed-form derivatives for stable gradient flow. Prior analysis\cite{murio1998discrete} confirms that this formulation achieves provably bounded error under discretization and measurement noise, making it well-suited for high-order and noise-prone inverse problems.

\underline{\textit{Formal definition:}} Drawing inspiration from Mollifiers\cite{Friedrichs_weakstrong} and the weak form formulation of PDEs (details in A.7), we construct a modified estimator, as illustrated in Fig. 1d. Although the base PhiML architecture remains unchanged aside from an added mollifier layer at the output, our methodology diverges from traditional PhiML approach in key ways:
\begin{itemize}
    \item Rather than estimating the target field \(\hat{u}\), the base model predicts \(\hat{g}\).
    \item \(\hat{u}\), is obtained at the Mollifying layer through the following operation (where $*$ is the convolution operator, $\eta$ is the Mollifying function and $U$ is the compact support of $\eta$):
    \begin{equation}
        \hat{u}(n) = \hat{g}*\eta(n) = \int_{m \in U} \hat{g}(m)\eta(n - m) \, dm,
    \end{equation} 
    \item At the Mollifier Layer, the higher-order derivatives w.r.t. a variable \(j\) (e.g., time or space) at a coordinate \(n\), such as \(\hat{u}_j\), \(\hat{u}_{jjj}\), etc., are obtained through the following operations:
    \begin{equation}
        \hat{u}_j = \hat{g}*\eta_j(n) = \int_{m \in U} \hat{g}(m) \eta_j(n - m) \, dm,
    \end{equation}
    \begin{equation}
        \hat{u}_{jjj} = \hat{g}*\eta_{jjj}(n) = \int_{m \in U} \hat{g}(m) \eta_{jjj}(n - m) \, dm.
    \end{equation}
\end{itemize}
This formulation offloads derivative computation to the analytic derivatives of a predefined mollifier $\eta$. Instead of recursive autodiff, we compute high-order derivatives by convolving the learned function $\hat{g}$ with $\eta_j$, $\eta_{jj}$, etc. \textit{This structural decoupling is the core innovation: it ensures efficient, stable, and noise-resilient high-order derivatives while sharply reducing memory and compute costs.} Detailed integration methodology is in Sec. A.4. The mollifier $\eta$ is designed to satisfy the following properties:

\begin{enumerate}
    \item \underline{Smoothness (Infinitely Differentiable)}: The mollifier $\eta$ is chosen from the $C^\infty$ class to enable computation of all high-order PDE derivatives.

   \item \underline{Compact Support}: $\eta$ has compact support on $U \subset \mathbb{R}^n$, meaning $\eta(m) = 0$ for all $m \notin U$. This confines integration to a finite region, reducing computational cost by restricting convolution to a fixed kernel size.
   \item \underline{Non-Negativity}: $\eta(m) \geq 0$ for all $m \in U$, ensuring the mollifier behaves as a localized averaging kernel and adheres to the following:
   \begin{enumerate}
   \item In the limit the support of $\eta$ contracts to zero, the mollifying operation converges to the identity operator, with $\eta$ approximating the Dirac delta $\delta(m)$, thereby establishing continuity with traditional PhiML formulations.
   \item Non-negativity prevents destructive interference during integration, mitigating cancellation errors common in oscillatory kernels.
   \item In physical systems with inherently non-negative targets (e.g., density, concentration), non-negative mollifiers preserve this constraint for consistent estimates.
\end{enumerate}
 
    \end{enumerate}

We experimented with various mollifying functions, including second-order polynomial, fourth-order polynomial, and sine function, within appropriate domains. Details on their form is in Sec. A.5.

With this formulation, mollifier layers provide the following key advantages over traditional autodiff-based physics-informed approaches: \textbf{(1) Computational efficiency:} Derivatives are computed via a single output-layer convolution, avoiding recursive autodiff and reducing memory usage. \textbf{(2) Structural decoupling:} Differentiation is shifted to fixed analytical kernels, independent of network depth. \textbf{(3) Noise robustness:} As localized smoothers, mollifiers suppress high-frequency artifacts, enabling stable inference under noise.

\underline{Convergence guarantees of the Mollifier layer.} [Uniform derivative consistency] Let the true field \(u\in C^{1}([0,1])\) be \(L\)-Lipschitz.  
Noisy grid samples are \(g_j=u(x_j)+n_j\) at \(x_j=jh\) with \(|n_j|\le\varepsilon\). Denote by \(\eta_\delta(r)=\delta^{-1}\eta(r/\delta)\) the analytic kernel from the mollifier layer and \(J_\delta g = \eta_\delta * g\). Then
\begin{equation}
  \bigl\|D_0(J_\delta g)-u'\bigr\|_\infty
  \;\le\;
  C_1\,\delta \;+\; C_2\,(h+\varepsilon),
\end{equation}
where \(D_0 g_j=(g_{j+1}-g_{j-1})/(2h)\). \noindent[Proof sketch.] Combine Murio \emph{et al.}\,\cite{murio1998discrete} Thm 2.6 (continuous bias) with Thm 3.5 (discrete \(+\) noise); see appendix A.8.  Choosing \(\delta\approx\sqrt{h}\) balances bias and variance, giving \(\mathcal{O}(\sqrt{h})\) uniform error while the backward operator norm remains \(\mathcal{O}(1/\delta)\).

\subsection{Inverse parameter estimation}
The proposed methodology estimates the spatio-temporally varying PDE parameter \(\hat{\lambda}(t, x)\). As discussed in Motivation \#2, even though this approach yields estimates for smoothly varying parameters; it may underperform when capturing high-frequency variations due to the inherent smoothness bias of neural network outputs. To mitigate this limitation, we introduce an alternative estimation strategy applicable when \(\lambda(t, x)\) can be expressed as a function of other predicted quantities—i.e., when \(\lambda\) is separable. For example, consider a PDE of the form $u_t-\lambda D[u]=0$, then our final estimated PDE parameter is $\hat{\lambda}_{final}=\frac{\hat{u}_t}{D[\hat{u}]}$. We propose that this estimation will better capture the ground truth, as \(\hat{u}\), optimized through the data loss (\(MSE_u\)), accurately reflects data variance and is better equipped to capture rapidly varying spatio-temporal features.

\begin{figure}
  \centering
  \includegraphics[width=\linewidth]{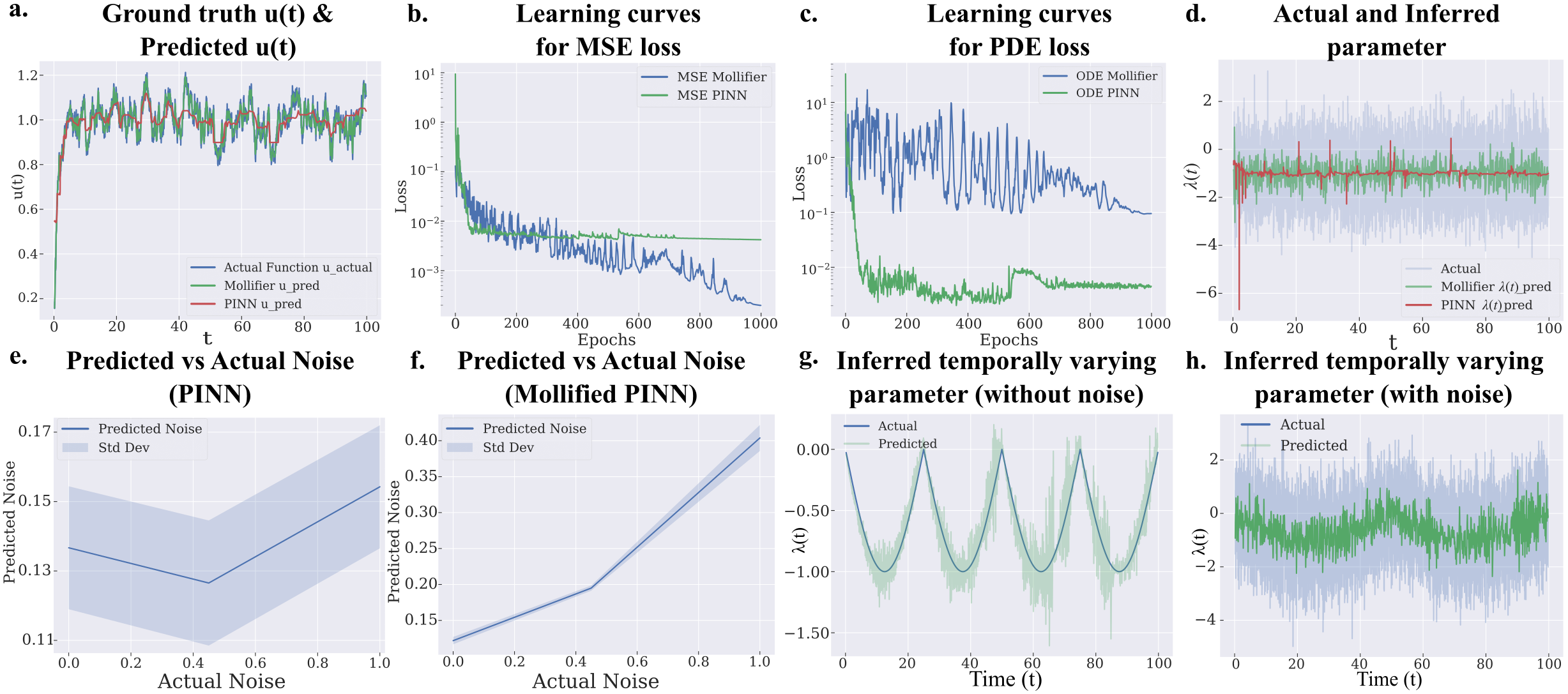}
  \caption{Parameter inference for the Langevin equation using PINN and Mollified PINN. a) Ground truth and predicted $u(t)$. b–c) $MSE_u$ Data and $MSE_f$ PDE residual learning curves. d) Ground truth and inferred forcing term $\lambda(t)$ under Gaussian noise. e–f) Actual vs. predicted noise trends in $\lambda(t)$. g) Inferred time-varying $\lambda(t)$ without noise. h) Inferred time-varying $\lambda(t)$ with added noise.}
\end{figure}

\section{Experiments}

We evaluate our method on three PDE benchmarks of increasing complexity: 1D Langevin (first-order), 2D heat (second-order), and 2D reaction–diffusion (fourth-order). We compare standard architectures—enhanced PINNs\cite{Raissi} (with residual connections\cite{Wang_gradflow}, spatio-temporal encodings\cite{wang2021eigenvector}, and layer norm\cite{ba2016layer}), PirateNet\cite{Wang_piratenet}, and PINNsFormer\cite{zhao2023pinnsformer}—against their mollified counterparts where applicable. These pairwise comparisons (native vs. native+mollifier) serve as direct ablation studies, isolating the impact of mollifier layers across models. The selected architectures reflect a spectrum of design choices: PINNs as the canonical baseline, PirateNet for its stability with higher-order derivatives, and PINNsFormer for its use of attention mechanisms\cite{vaswani2017attention, dosovitskiy2020image, niu2021review}. Due to the high computational cost of attention-based models, we do not include a mollified PINNsFormer; mollified PINNs already achieve superior performance at a fraction of the computational cost. While all quantitative comparisons are reported in the main text, we plot only PINN-based variants for clarity. Additional plots are provided in the appendix.

\subsection{First-Order 1D Langevin equation}

To evaluate parameter inference under time-varying dynamics, we consider a simplified form of the Langevin equation—a classical model for systems subject to both deterministic and stochastic forces. Specifically, we use the rescaled first-order ODE:
\begin{equation}
u_t = u + \lambda(t),
\end{equation}
where $u(t)$ is the observed velocity and $\lambda(t)$ is an unknown forcing term to be inferred. This formulation preserves the essential Langevin‐dynamics structure—modeling input‐driven transients—while sidestepping second‐order stochastic complexity. As our simplest nontrivial ODE, it offers a clear benchmark for evaluating inference accuracy and robustness. Under constant forcing $\lambda$, all models perform similarly in accuracy and speed (see B.1; Fig. 5–7).

\subsubsection{Langevin equation with Gaussian White Noise}

We simulated the Langevin equation with a forcing term of fixed mean ($\Lambda = -1$) and varying noise levels ($\sigma \in \{0, 0.44, 1\}$) to benchmark all models. For consistency, we set $N_u = N_f$ across all experiments, using a original mollifying function (Sec. A.5) as $\eta$. As shown in Fig. 2a–c for $\sigma = 1$, both PINNs and mollified PINNs converge to the underlying function, but the latter—combined with our parameter estimation scheme—more accurately captures noise variations (Fig. 2d). Across noise levels, mollified PINNs consistently outperform standard PINNs in capturing these trends (Fig. 2e). Similar trends are also obtained for PirateNet and PINNsFormer, where the mollified versions surpass native architectures (Fig. 8-10).

We evaluated each method’s accuracy—via mean and variance correlation—and computational efficiency—via training time and memory—over five runs per noise level. Mollified variants consistently outperformed their native counterparts in tracking temporal trends and used fewer resources on average (Table 2; see B.4 for details). Kernel‐type ablations in Appendix B.1.3 (Figs. 11–12) further show that mollifier shape and support critically affect noise handling and inference accuracy.

\subsubsection{Langevin Equation with temporally varying force}
We next evaluate the performance of the models in inferring a noiseless, temporally varying parameter. For this, we simulate a trajectory with a time-dependent mean, \(\lambda(t) \sim \mathcal{N}(\Lambda(t), 0)\), and generate simulations for various configurations of \(\Lambda(t)\) to compare the capabilities of the native models with their mollified versions. As shown in Fig. 2g, mollified PINNs provide a very good estimate of the underlying parameter's variation. While all variants successfully capture the global mean, native variants generally fail to capture the temporal variation in the parameter Fig. 13-16.

We introduced noise (\(\sigma = 1\)) to the underlying parameters. As illustrated in Fig. 2h, mollified PINNs demonstrate great performance in capturing temporal trends in the forcing term compared to PINNs Fig 13. Similar trends are obtained for the other models Fig . 14-16, proving that the mollifier layer lends higher accuracy predictions.

\begin{figure}
  \centering
  \includegraphics[width=\linewidth]{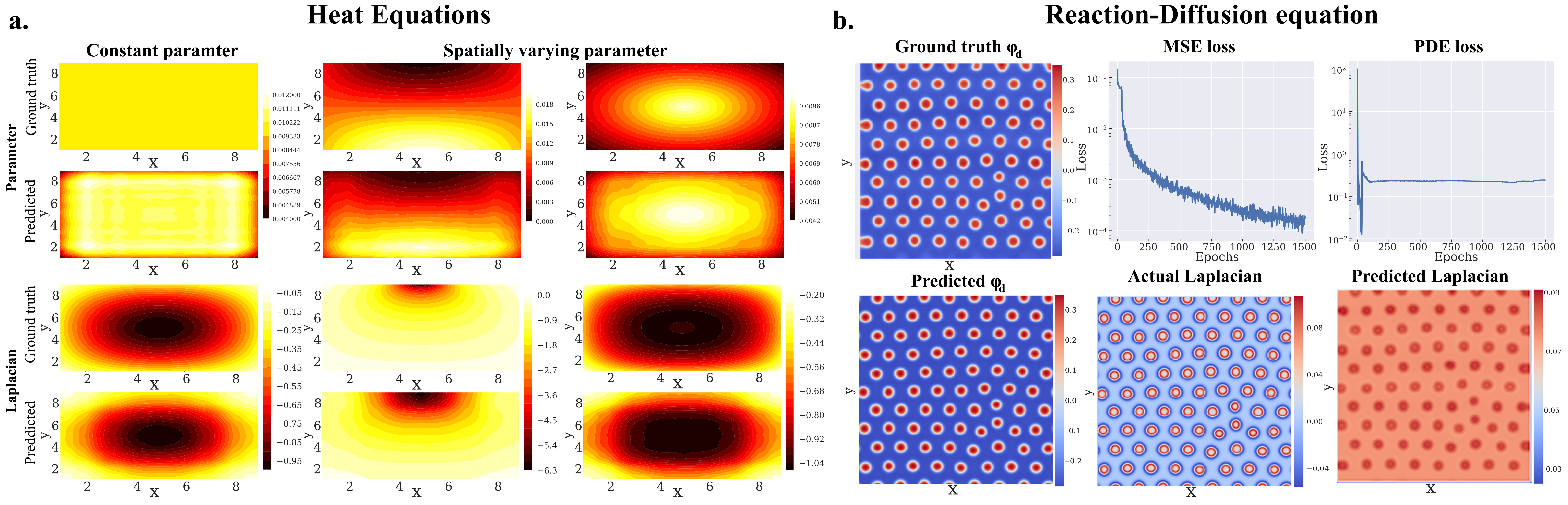}
  \caption{Parameter estimation for the Heat and Reaction–Diffusion equations using Mollified PINN. a) Ground truth and predictions for constant and spatially varying thermal diffusivity $\lambda(x, y)$ and Laplacians for the 2D heat equation. b) Predicted vs. actual chromatin order parameter $\phi_d$, recovered Laplacian, and training curves for the reaction–diffusion system.}
\end{figure}

\subsection{Second-Order 2D Heat Equation}

We consider the steady-state 2D heat equation, a second-order PDE modeling spatial temperature distribution under thermal diffusion and external sources:

\begin{equation}
    0 = \lambda(x, y) \nabla^2 u + m(x, y),
\end{equation}

where $u(x, y)$ is the observed temperature field, $\lambda(x, y)$ is the unknown thermal diffusivity, and $m(x, y)$ is a known source term. The goal is to recover $\lambda(x, y)$ from a sparsely sampled $u(x, y)$ and a known $m$, a problem relevant in materials science, geophysics, and biomedical imaging for characterizing heterogeneous media.

We evaluate all baseline models and their mollified counterparts on cases with both constant and spatially varying diffusivity (Fig. 3a shows results for mollified PINNs; others are in the appendix). Mollified models consistently outperform their native versions in capturing spatial trends while significantly reducing training time and memory usage (Fig. 17-20). Quantitative results, including correlation metrics and efficiency comparisons, are summarized in Table 2.

\subsection{Fourth-Order 2D Reaction–Diffusion Equation}

Reaction–diffusion systems model spatial pattern formation via diffusion and local reaction kinetics, with applications in biology, chemistry, and ecology\cite{Epstein_reac_diff}. We study a biologically motivated system describing DNA organization in the eukaryotic/human nucleus, where heterochromatin (inactive) and euchromatin (active) states are shaped by diffusion and epigenetic reactions such as methylation and acetylation—key drivers of gene regulation\cite{Kant,vinayak2025polymer}. Our goal is to infer the spatially varying reaction rate $\lambda(x, y)$ from the final chromatin configuration in synthetically generated data, enabling scalable estimation of biophysical parameters. In Sec. 5, we extend this framework to experimentally obtained super-resolution imaging data.

We adopt a phase-field model\cite{Kant} with volume fractions $\phi_h$, $\phi_e$, and $\phi_n$ for heterochromatin, euchromatin, and nucleoplasm, respectively. The system is governed by the order parameter $\phi_d = \phi_h - \phi_e$, whose dynamics follow:

\begin{equation}
    \frac{\partial \phi_d}{\partial t} = \nabla^2 \mu_d + 2(\lambda \phi_e - \phi_h),
\end{equation}

where $\mu_d$ is a chemical potential depending on $\phi_h$ and $\phi_e$. At steady state, this yields:

\begin{equation}
    \nabla^2 \mu_d + 2(\lambda \phi_e - \phi_h) = 0.
\end{equation}

Since $\mu_d$ includes second-order derivatives of $\phi_d$, i.e.,  $\mu_d\propto\nabla^2\phi_d$, the system effectively forms a fourth-order PDE, making it a challenging testbed for high-order derivative learning (Sec. B.3.1).

As shown in Fig. 3b, mollified PINNs accurately estimate $\lambda(x, y)$ and stably recover the Laplacian—an essential intermediate—unlike standard PINNs, which often misestimate its magnitude (see also Fig. 1). Mollified models consistently outperform their native counterparts in both accuracy and efficiency (Table 2, Fig. 21-24).

Fig. 4a further shows that mollified PINNs capture spatial noise trends in $\lambda(x, y)$, which the native PINNs fail to achieve demonstrating effectiveness in capturing heterogeneity (plots for other models provided in Fig. 21-24). These results underscore the effectiveness of mollifier layers in high-order, noise-sensitive inverse problems in biophysics.

\begin{table}[t]
\centering
  \caption{
    Comparison of baseline and mollified models across PDEs and metrics. 
    \textbf{PINN}: Physics-Informed Neural Network, 
    \textbf{PN}: PirateNet, 
    \textbf{+M} indicates the corresponding model augmented with mollifier layers.
  }
  \label{tab:structured-table}
  \centering
  \resizebox{\textwidth}{!}{%
  \begin{tabular}{l@{\hskip 8pt}c@{\hskip 8pt}c@{\hskip 6pt}c@{\hskip 6pt}c@{\hskip 6pt}c@{\hskip 6pt}c}
    \toprule
    \textbf{PDE constraint} & \textbf{Metric} & \textbf{PINN} & \textbf{+M} & \textbf{PN} & \textbf{PN+M} & \textbf{PINNsFormer} \\
    \cmidrule(lr){3-4} \cmidrule(lr){5-6} \cmidrule(lr){7-7}
    
    \multirow{5}{*}{\textbf{1D Langevin (1\textsuperscript{st})}} 
        & \textbf{Time Taken (sec)}      & 2138  & \textbf{1615}  & 2414 & \textbf{1473} & 6507 \\
        & \textbf{Mean Corr.}      & \textbf{0.99} & 0.96  & 0.98 & \textbf{0.99} & 0.98 \\
        & \textbf{Temporal Corr.}   & \textbf{0.99} & 0.97 & \textbf{0.98} & 0.97 & 0.35 \\
        & \textbf{Peak Memory (GB)}     & 0.21  &  \textbf{0.16} & 0.77 & \textbf{0.11}  & 0.33 \\
    \addlinespace
    
    \multirow{5}{*}{\textbf{2D Heat (2\textsuperscript{nd})}} 
        & \textbf{Time Taken (sec)}      & 2294  & \textbf{1582} & 4737 & \textbf{2948} & 10200 \\
        & \textbf{Mean Corr.}      & 0.81  & \textbf{0.99}  & 0.13 & \textbf{0.22} & 0.73 \\
        & \textbf{Spatial Corr.} & 0.21  & \textbf{0.99}  & 0.16 & \textbf{0.20} & 0.04 \\
        & \textbf{Peak Memory (GB)}     & 1.20  &  \textbf{0.24} & 1.15 & \textbf{0.11} & 0.8  \\
        & \textbf{Laplacian Corr.}  & 0.16 & \textbf{0.99} & 0.01 & \textbf{0.23} & 0.09 \\
    \addlinespace
    
    \multirow{5}{*}{\textbf{2D Reaction-Diffusion (4\textsuperscript{th})}} 
        & \textbf{Time Taken (sec)}      & 3386 & \textbf{335} & 817 & \textbf{125} & 34487 \\
        & \textbf{Mean Corr.}      & 0.44 & \textbf{0.99} & 0.77 & \textbf{0.99} & 0.68 \\
        & \textbf{Spatial Corr.}   & 0.17 & \textbf{0.84} & 0.25 & \textbf{0.91} & 0.47 \\
        & \textbf{Peak Memory (GB)}     & 2.75 & \textbf{0.23} & 0.48 & \textbf{0.12} & 1.90 \\
        & \textbf{Laplacian Corr.}  & 0.21 & \textbf{0.78} & 0.01 & \textbf{0.75} & 0.01 \\
    \bottomrule
  \end{tabular}
  }
\end{table}
\section{Biophysical Application}

DNA architecture spans multiple spatial scales, with nanoscale features such as "heterochromatin domains" lying below the diffraction limit of conventional microscopy\cite{ricci2015chromatin}. Super-resolution techniques\cite{schermelleh2019super} like STORM\cite{rust2006stochastic} overcome this barrier, enabling visualization of sub-diffraction chromatin structures and their spatial heterogeneity (Fig. 4b). While such imaging provides rich qualitative insight, extracting mechanistic information demands integration with physical models.

We use mollified PINNs to infer spatially varying epigenetic reaction rates from STORM images, selecting PINNs for their simplicity and training efficiency. These rates parameterize a reaction–diffusion model (discussed in Sec. 4.3) that captures the interplay between biochemical transformations (e.g., methylation) and chromatin mobility. Each $1.5 \times 1.5\,\mu\text{m}^2$ image region is preprocessed into a continuous chromatin density field using Watson kernel smoothing (see Appendix C). Mollified PINNs then infer local reaction rates by minimizing the residual of the governing PDE (eq. 11), computing high-order derivatives via analytic mollifier convolutions.

As shown in Fig. 4c, mollified PINNs recover both the mean and variance of reaction rates with high fidelity, aligned with ground-truth chromatin domain structure and noise, benchmarked in the previous section using simulations (Sec. 4.3). Resolving local epigenetic reaction rates from imaging reveals the mechanistic drivers of chromatin reorganization, linking nanoscale domain remodeling to gene regulation, cancer metastasis, and cell fate memory in development and disease contexts\cite{vinayak2025polymer}. Our results thus establish mollifier layers as a robust tool for making PhiML scalable for integrating super-resolution imaging with biophysical modeling to extract interpretable parameters from high-dimensional biological data.

\begin{figure}
  \centering
  \includegraphics[width=\linewidth]{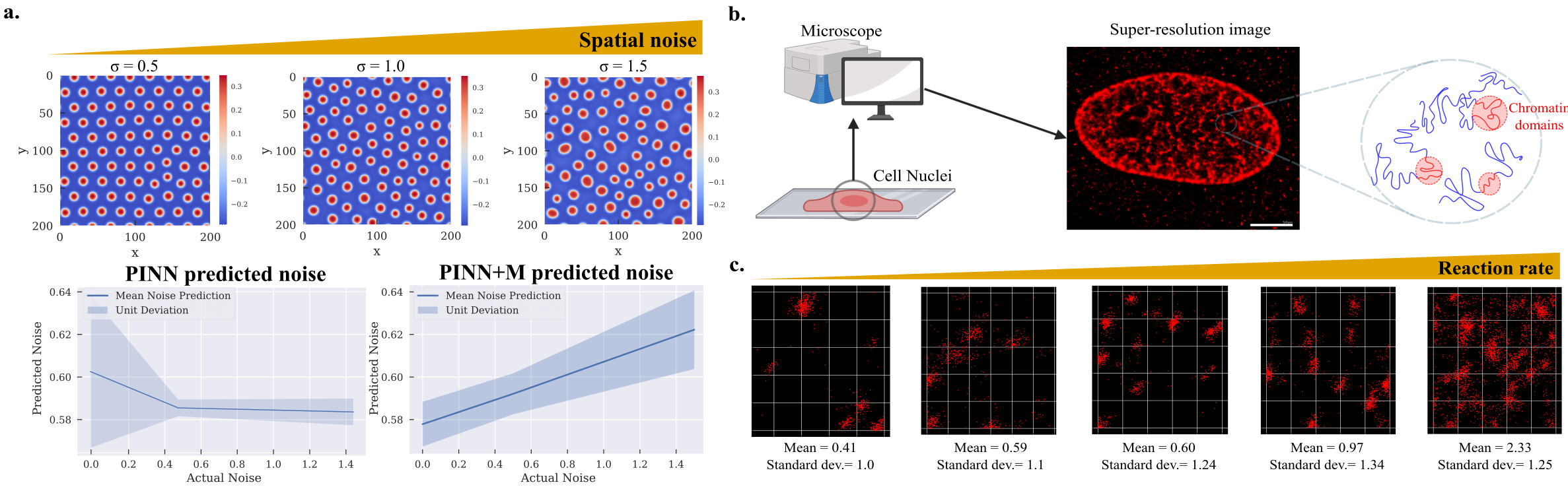}
  \caption{Mollified PINN capture spatial noise trends and extract reaction rate statistics in the DNA Reaction–Diffusion system. a) Spatial noise trends in inferred reaction rates from synthetic data. b,c) Predicted mean and variance of reaction rates from super-resolution images of human nuclei.}

\end{figure}

\section{Discussion}
By collapsing recursive higher-order automatic differentiation into a single analytic convolution, Mollifier Layers reduce both memory footprint and training time by 6–10×, as demonstrated in our experiments, while preserving robust, high-order derivative estimates. Though tested for inverse PDEs, the mollifier-based derivative layer extends naturally to forward solvers (Sec. B.5), operator learning, and neural ODEs where stable and accurate gradients are essential. This efficiency enables scalable physics-informed machine learning to tackle complex systems like kilometer-scale weather models\cite{bodnar2024aurora,palmer2019stochastic} or morphogenesis\cite{liu2024morphogenesis,wyczalkowski2012computational} simulations with markedly lower computational cost.

Nonetheless, performance remains sensitive to the choice of mollifier kernel (see Sec. B.1.3), which must trade off noise suppression against bias. The current hand-tuned implementation is limited near boundaries and on anisotropic grids. Developing adaptive or learned kernels, boundary-aware formulations, and validation strategies for adaptive meshes are key directions for future work.

\section{Related Works}

Recent physics-informed machine learning (PhiML) approaches—ranging from PINNs\cite{Raissi} and DeepONets\cite{Lu_FNO} to FNOs\cite{Li_FNO}, PINOs\cite{li2024physics}, and PIKANs\cite{patra2024physics}—have advanced PDE modeling. Our focus is on inverse problems and stable derivative computation, replacing recursive autodiff with a convolutional alternative. We review related work in these directions and outline our contributions.

\underline{Inverse Modeling:} Recent work mitigates inverse‑PhiML shortcomings in complementary ways. PINNverse\cite{almanstotter2025pinnverse} treats parameter recovery as a constrained optimization, improving robustness to noise; H‑PINN\cite{chandrasukmana2025hadamard} enforces Hadamard well‑posedness to regularize ill‑posed tasks; Bayesian PINNs\cite{yang2021b} place probabilistic priors on parameters to quantify uncertainty, albeit with significant sampling cost; and gPINNs\cite{yu2022gradient} augment the loss with gradient information to accelerate convergence; The PI-INN framework \cite{guan2024efficient} enables Bayesian inference via invertible networks but retain the full autodiff graph. Beyond the PINN family, operator‑learning models such as the Latent Neural Operator\cite{wang2024latent} and Transformer‑BVIP\cite{guo2022transformer} use cross‑attention or boundary‑aware transformers for inverse PDEs. Despite these advances, all rely on recursive autodiff. We replace it with a differentiable mollifier layer, enabling stable, memory-efficient, high-order derivatives while remaining plug-and-play.

\underline{Alternatives to Autodiff:} Prior efforts to avoid recursive autodiff fall into three tracks: (i) finite‑difference PhiML\cite{lim2022physics,chiu2022can,huang2024efficient} that save memory but lose high‑order accuracy and suffers at high resolution, (ii) spectral PhiML\cite{maust2022fourier,patel2022thermodynamically} that gain stability yet demand bespoke bases and (iii) operator‑specific networks—e.g., Koopman operator nets\cite{lusch2018deep}, implicit Fourier/Green‑function networks\cite{li2024neural}—that embed analytical derivative kernels directly in the architecture; these deliver fast inference for a fixed PDE class but cannot generalize beyond the operators they hard‑code. In contrast, our approach offers a general, architecture-agnostic alternative that enables stable, low-memory, high-order gradients—even in noisy, fourth-order PDEs—without sacrificing flexibility across PDE classes.

\section{Conclusion}

Mollifier Layers offer a robust, scalable alternative to recursive autodiff, enabling stable high-order derivative estimation critical for inverse PDE learning. By decoupling differentiation from network depth, they unlock accurate, noise-resilient inference across complex physical systems. In doing so, Mollifier Layers pave the way for physics-informed machine learning to tackle previously intractable regimes—where deep models must reason stably over space, time, and noise—to extract scientific insight from data in real-world dynamical systems.

\section*{Acknowledgements}

This work was supported by NIH Award U54CA261694 (V.B.S.); NCI Awards R01CA232256 (V.B.S.); NSF CEMB Grant CMMI-154857 (V.B.S.); NSF Grants MRSEC/DMR-1720530 and DMS-1953572 (V.B.S.); and NIBIB Awards R01EB017753 and R01EB030876 (V.B.S.).

\printbibliography

\appendix

\section{Details of the models}
\subsection{Details of the PINNs implementation}

\begin{figure}[ht]
\vskip 0.2in
\begin{center}
\centerline{\includegraphics[width=\columnwidth]{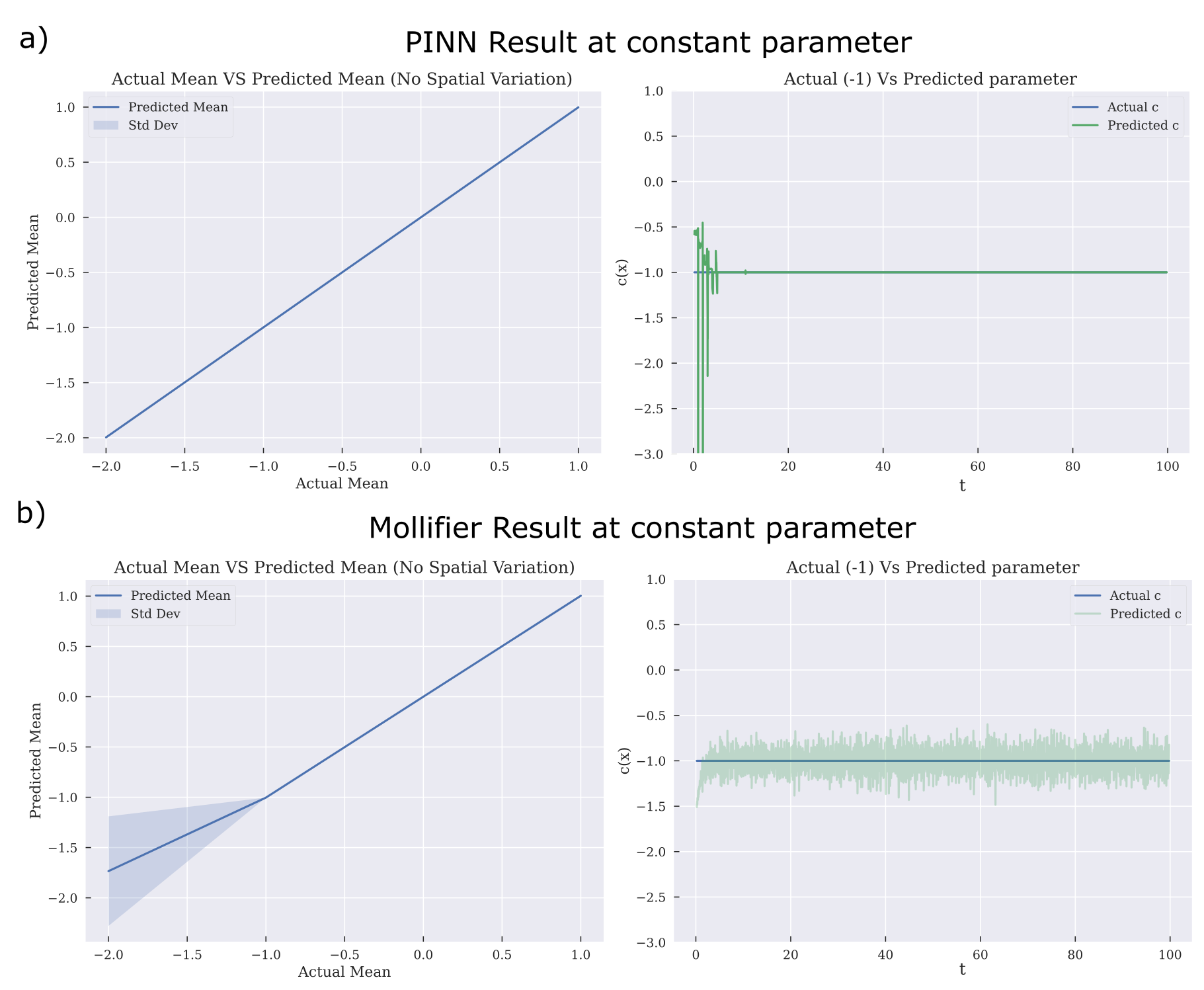}}
\caption{\textit{Constant forcing term inference in Langevin equation using PINN and Mollified PINN}}
\label{icml-historical}
\end{center}
\vskip -0.2in
\end{figure}

\begin{figure}[ht]
\vskip 0.2in
\begin{center}
\centerline{\includegraphics[width=\columnwidth]{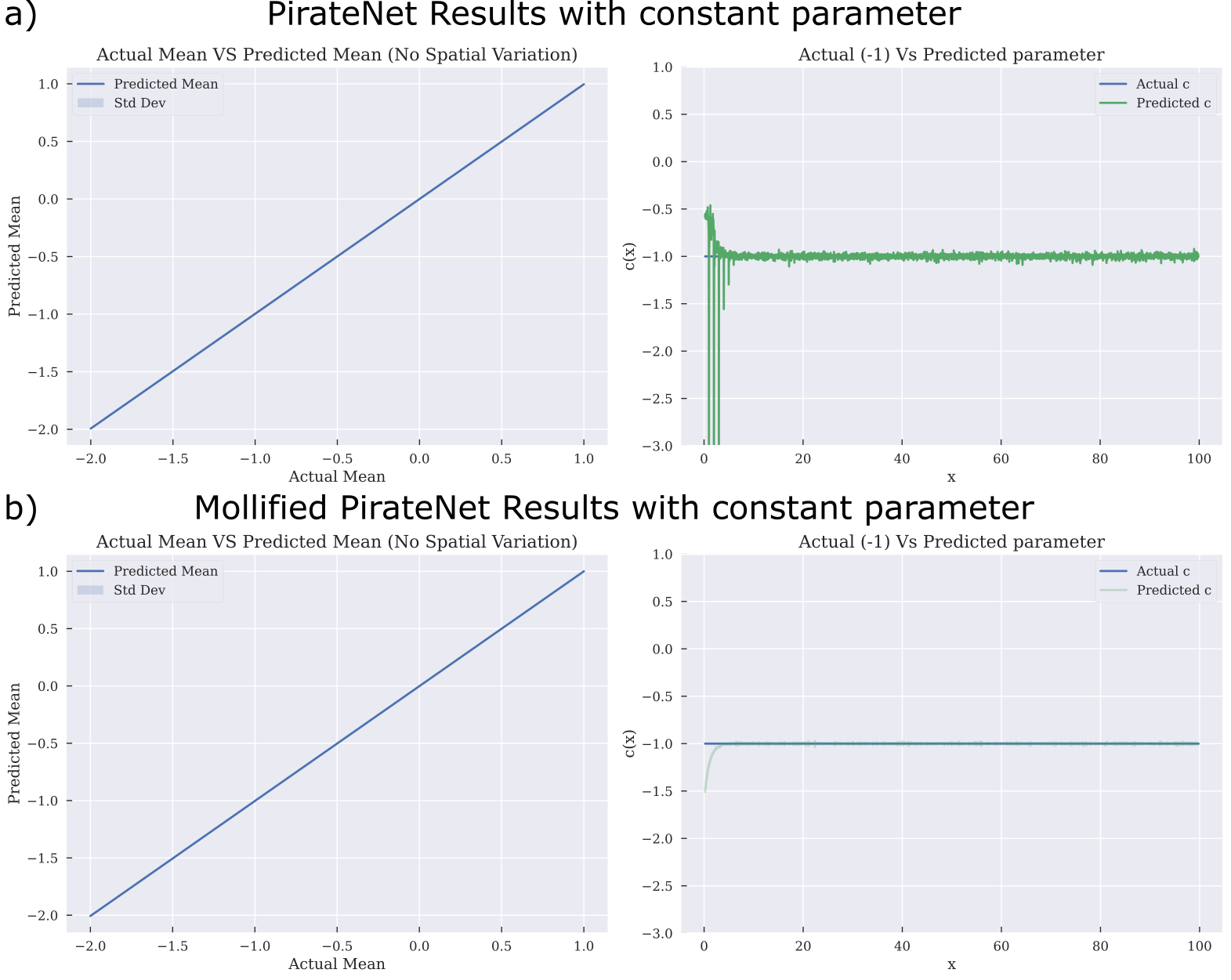}}
\caption{\textit{Constant forcing term inference in Langevin equation using PirateNet and Mollified PirateNet}}
\label{icml-historical}
\end{center}
\vskip -0.2in
\end{figure}

\begin{figure}[ht]
\vskip 0.2in
\begin{center}
\centerline{\includegraphics[width=\columnwidth]{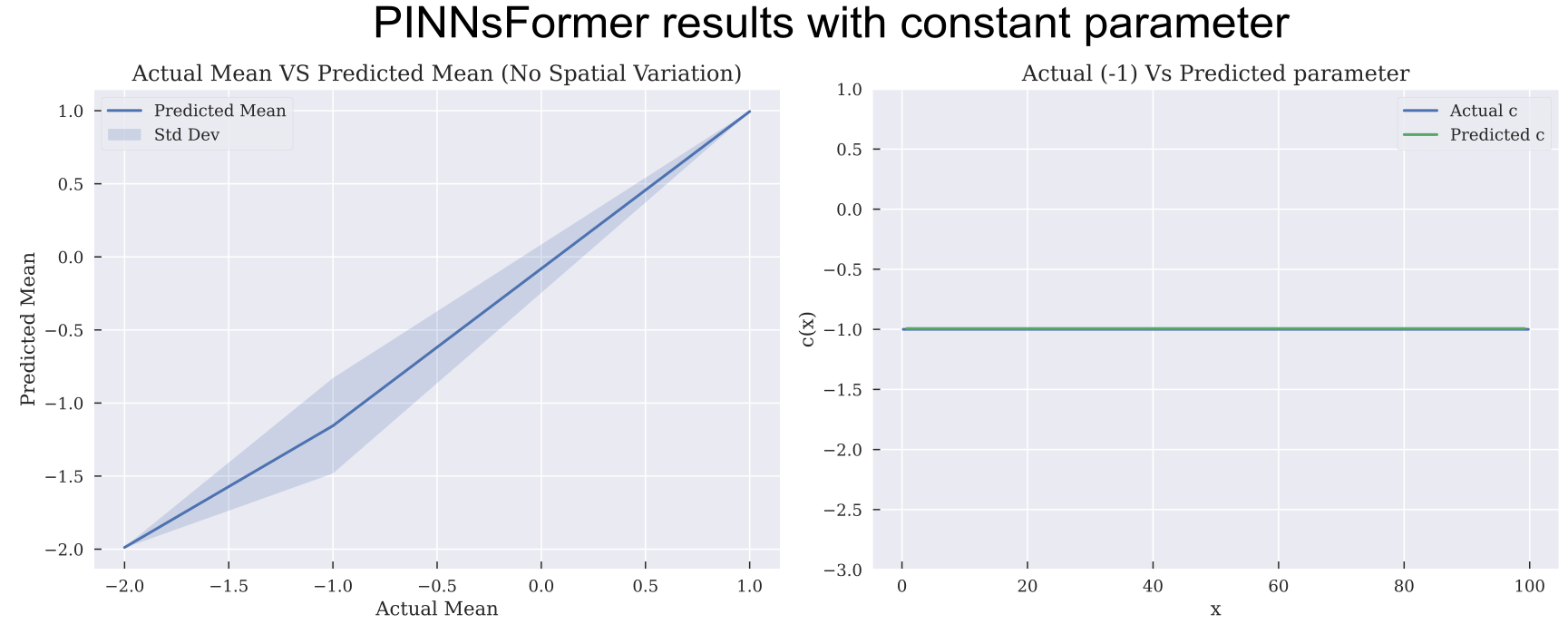}}
\caption{\textit{Constant forcing term inference using PINNsFormer}}
\label{icml-historical}
\end{center}
\vskip -0.2in
\end{figure}

\begin{figure}[ht]
\vskip 0.2in
\begin{center}
\centerline{\includegraphics[width=\columnwidth]{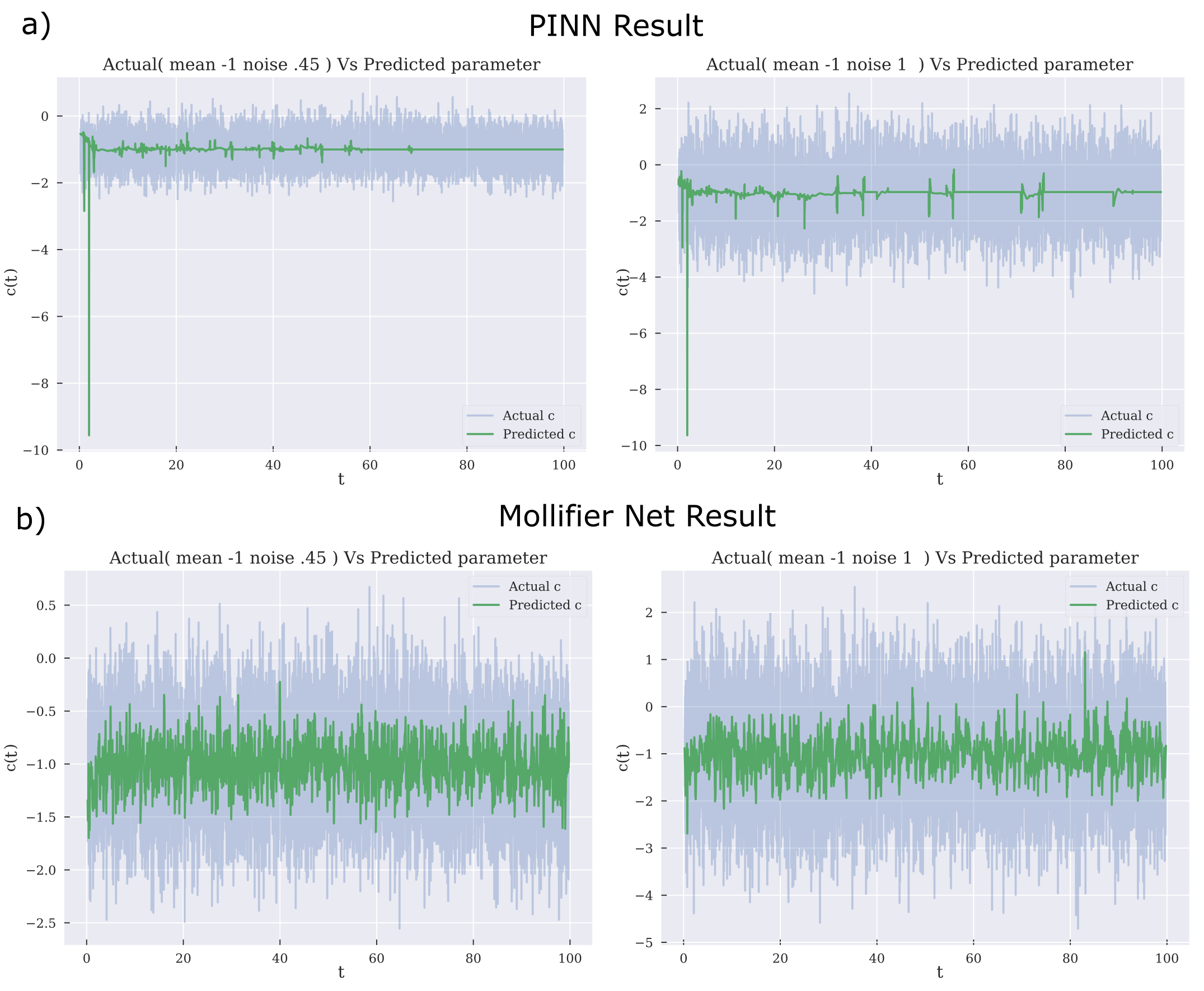}}
\caption{\textit{Noisy forcing term inference in Langevin equation using PINN and Mollified PINN}}
\label{icml-historical}
\end{center}
\vskip -0.2in
\end{figure}

\begin{figure}[ht]
\vskip 0.2in
\begin{center}
\centerline{\includegraphics[width=\columnwidth]{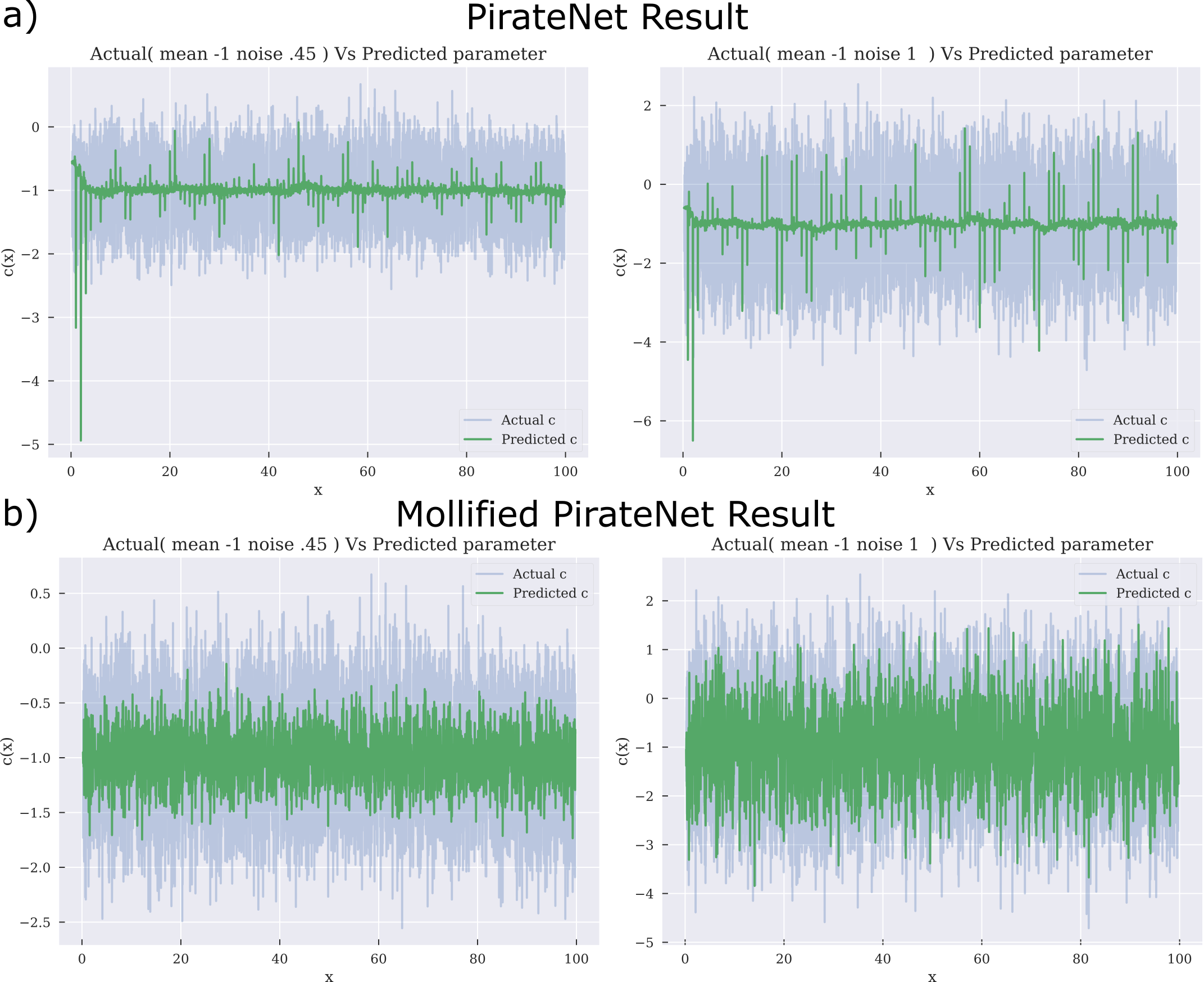}}
\caption{\textit{Noisy forcing term inference in Langevin equation using PirateNet and Mollified PirateNet}}
\label{icml-historical}
\end{center}
\vskip -0.2in
\end{figure}

\begin{figure}[ht]
\vskip 0.2in
\begin{center}
\centerline{\includegraphics[width=\columnwidth]{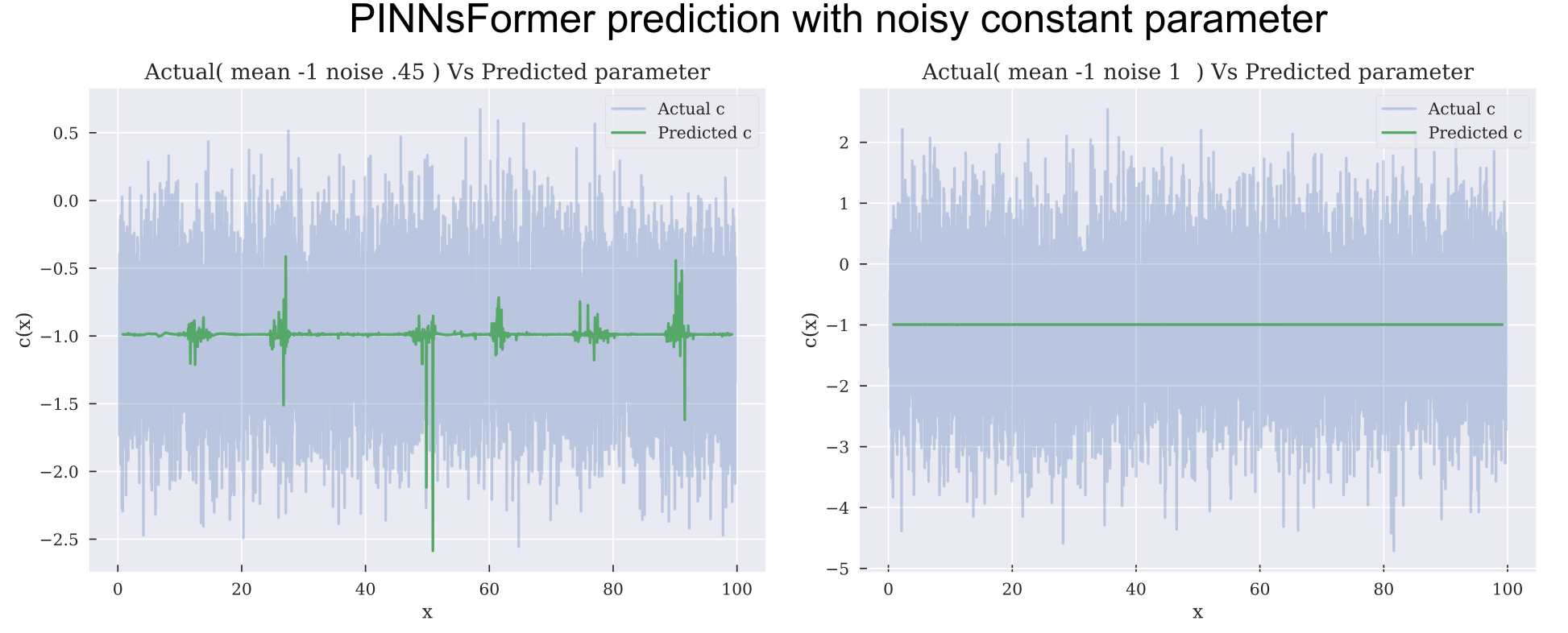}}
\caption{\textit{Noisy forcing term inference in Langevin equation using PINNsFormer}}
\label{icml-historical}
\end{center}
\vskip -0.2in
\end{figure}

\begin{figure}[h]
\vskip 0.2in
\begin{center}
\centerline{\includegraphics[width=\columnwidth]{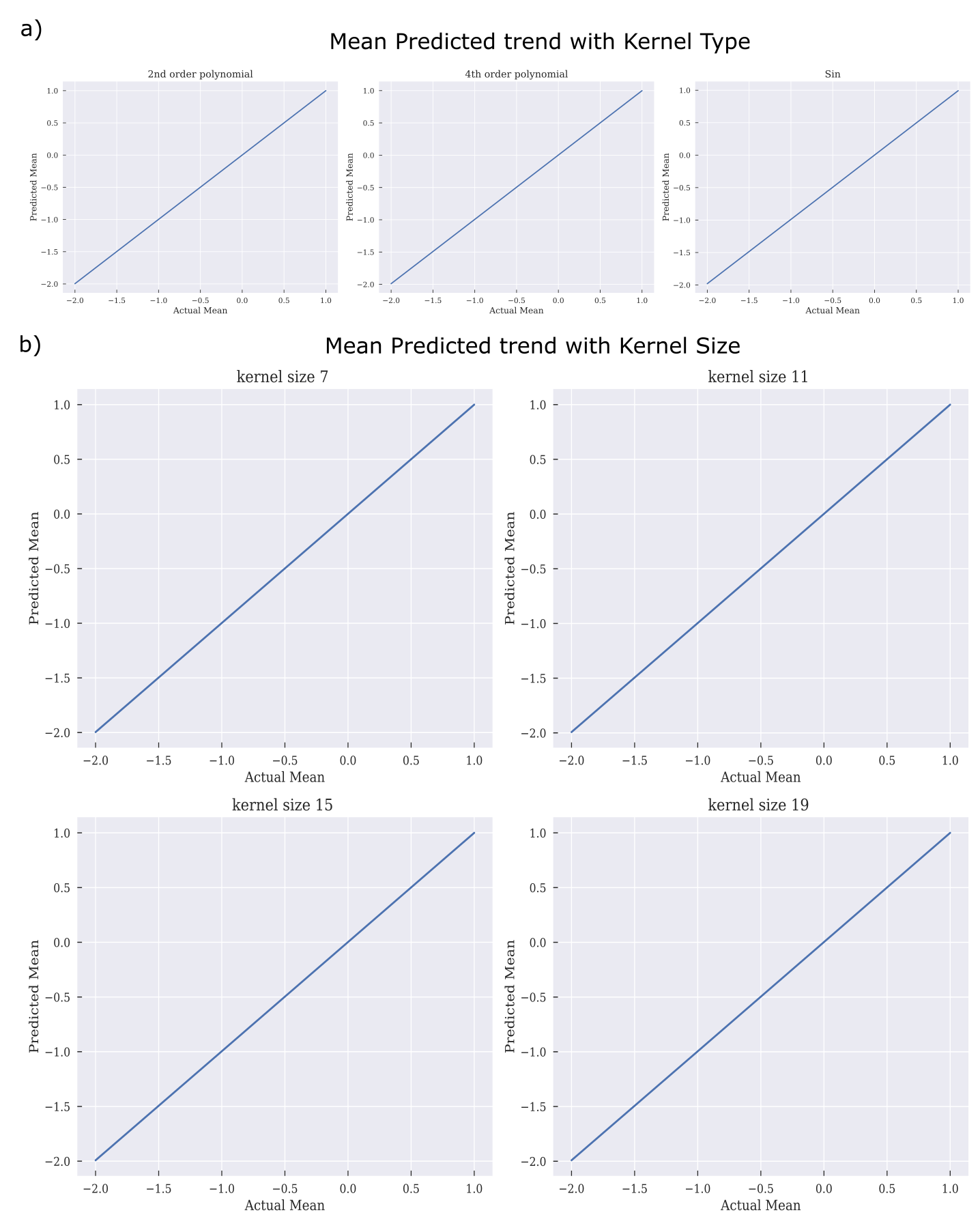}}
\caption{\textit{Mean Forcing term prediction with different kernel choices with Mollified PINNs}}
\label{icml-historical}
\end{center}
\vskip -0.2in
\end{figure}

\begin{figure}[h]
\vskip 0.2in
\begin{center}
\centerline{\includegraphics[width=\columnwidth]{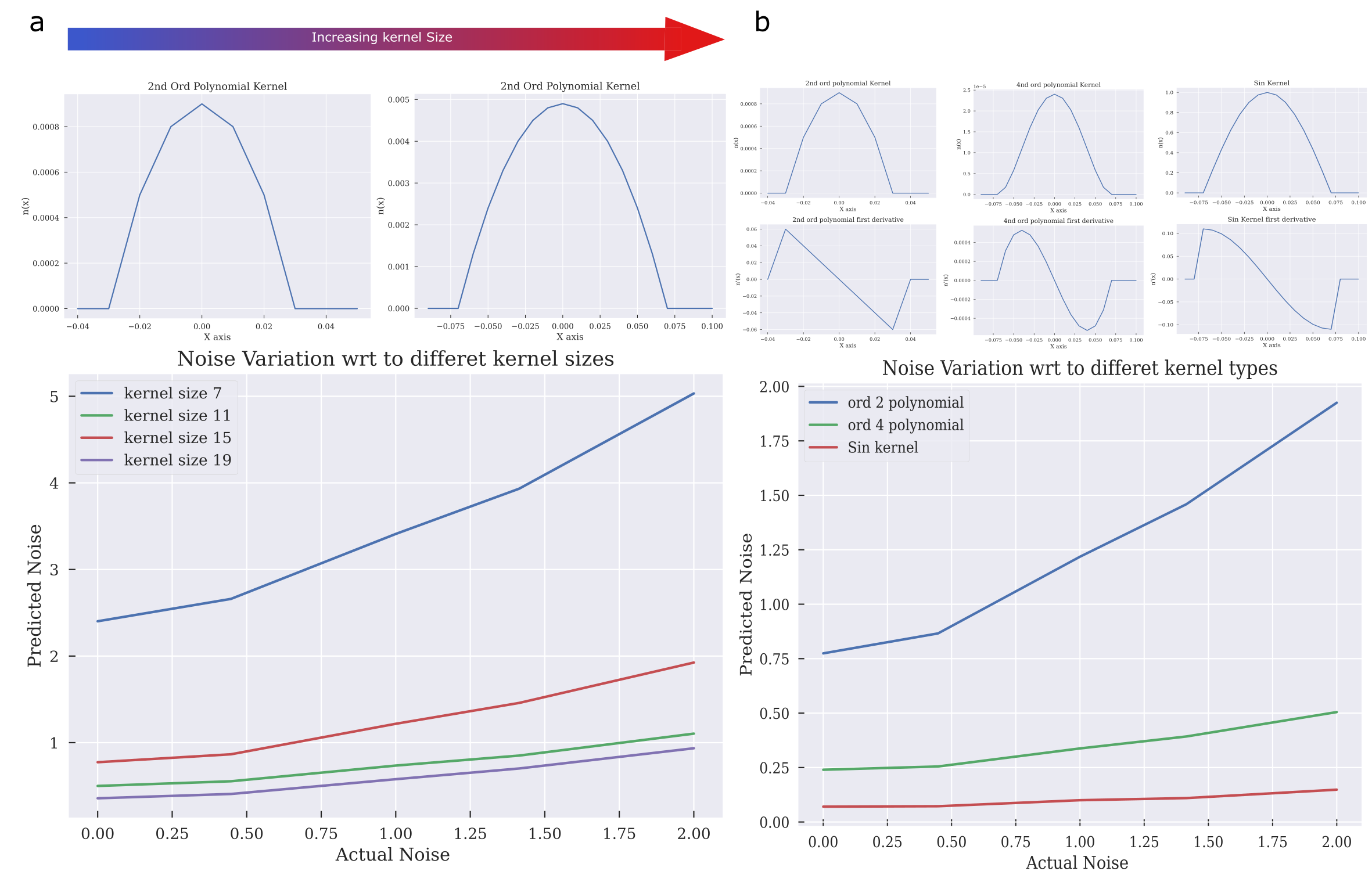}}
\caption{\textit{Estimating the Noise in the forcing term prediction with different kernel choices with Mollified PINNs}}
\label{icml-historical}
\end{center}
\vskip -0.2in
\end{figure}

\begin{figure}[h]
\vskip 0.2in
\begin{center}
\centerline{\includegraphics[width=\columnwidth]{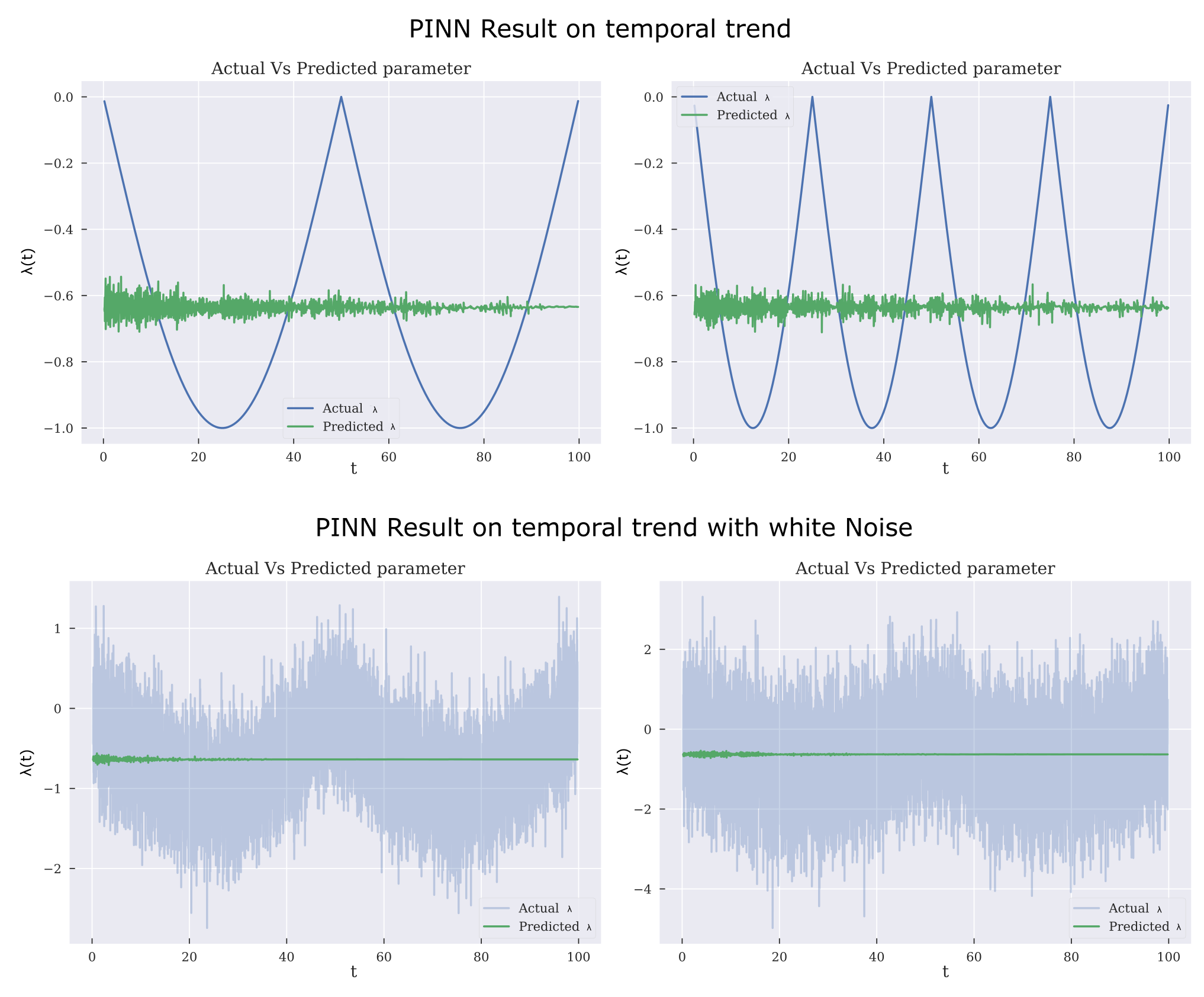}}
\caption{\textit{Estimating with and without noise temporally varying forcing term with PINNs}}
\label{icml-historical}
\end{center}
\vskip -0.2in
\end{figure}

\begin{figure}[h]
\vskip 0.2in
\begin{center}
\centerline{\includegraphics[width=\columnwidth]{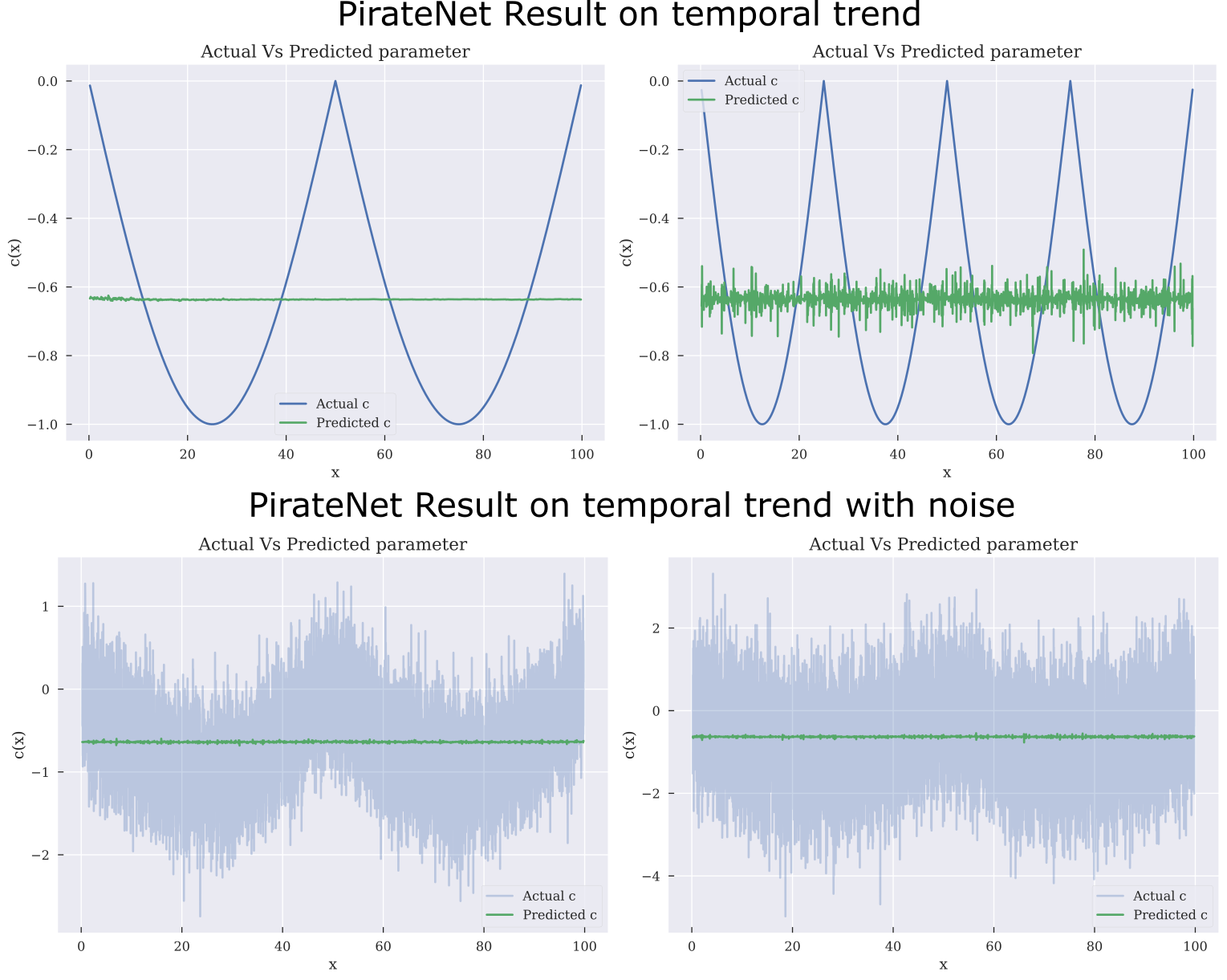}}
\caption{\textit{Estimating with and without noise temporally varying forcing term with PirateNet}}
\label{icml-historical}
\end{center}
\vskip -0.2in
\end{figure}

\begin{figure}[h]
\vskip 0.2in
\begin{center}
\centerline{\includegraphics[width=\columnwidth]{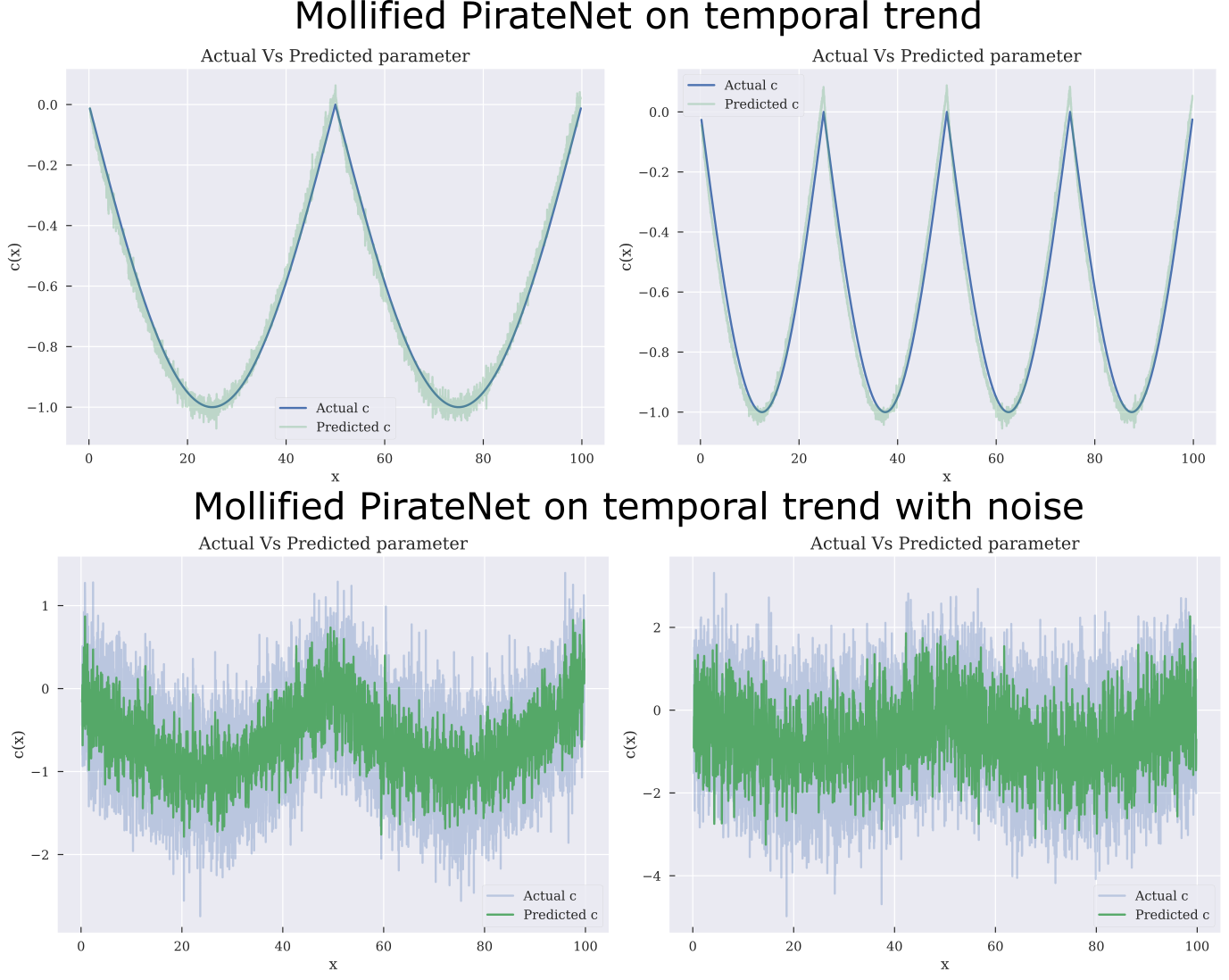}}
\caption{\textit{Estimating with and without noise temporally varying forcing term with Mollified PirateNet}}
\label{icml-historical}
\end{center}
\vskip -0.2in
\end{figure}

\begin{figure}[h]
\vskip 0.2in
\begin{center}
\centerline{\includegraphics[width=\columnwidth]{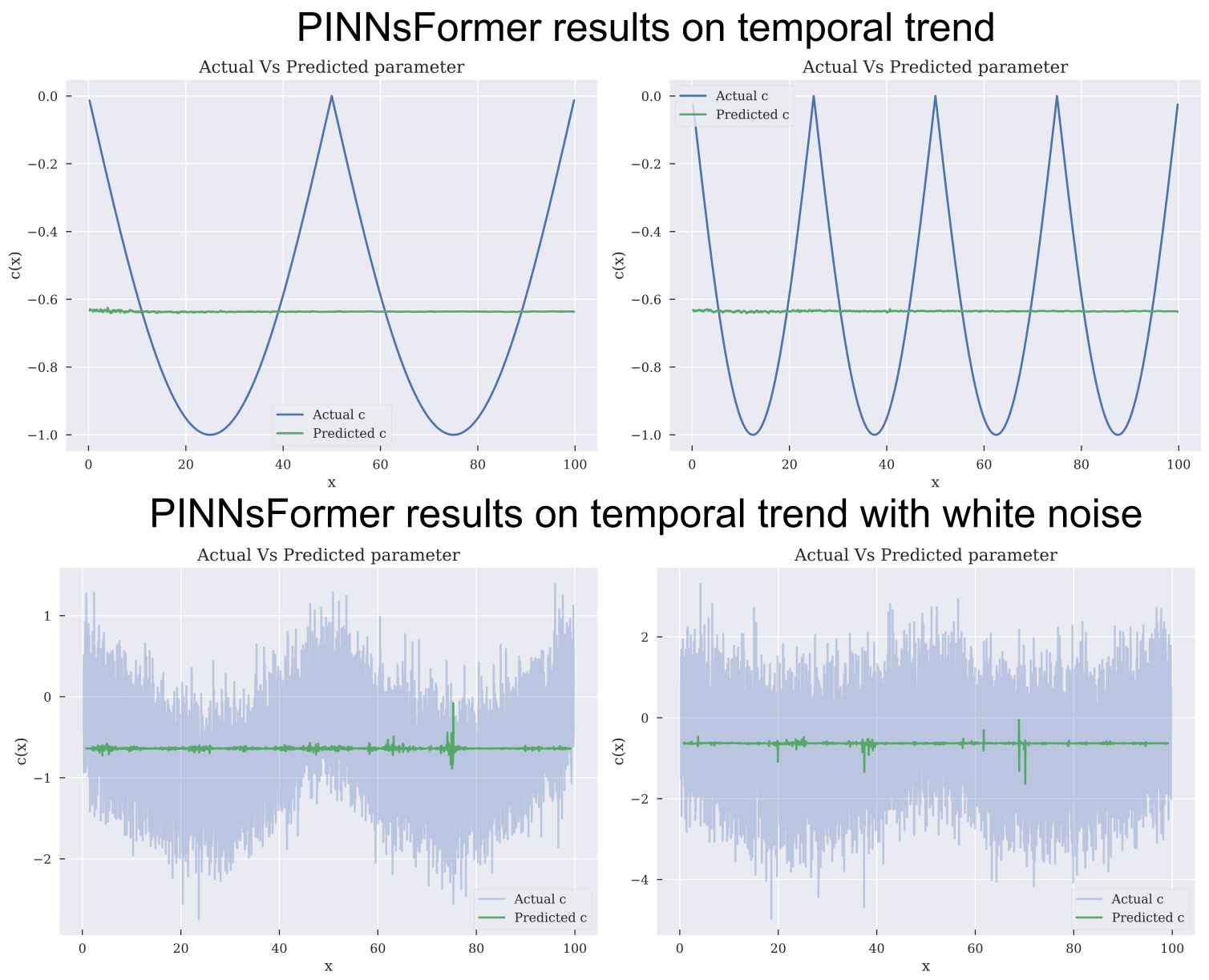}}
\caption{\textit{Estimating with and without noise temporally varying forcing term with PINNsFormer}}
\label{icml-historical}
\end{center}
\vskip -0.2in
\end{figure}

\begin{figure}[h]
\vskip 0.2in
\begin{center}
\centerline{\includegraphics[width=\columnwidth]{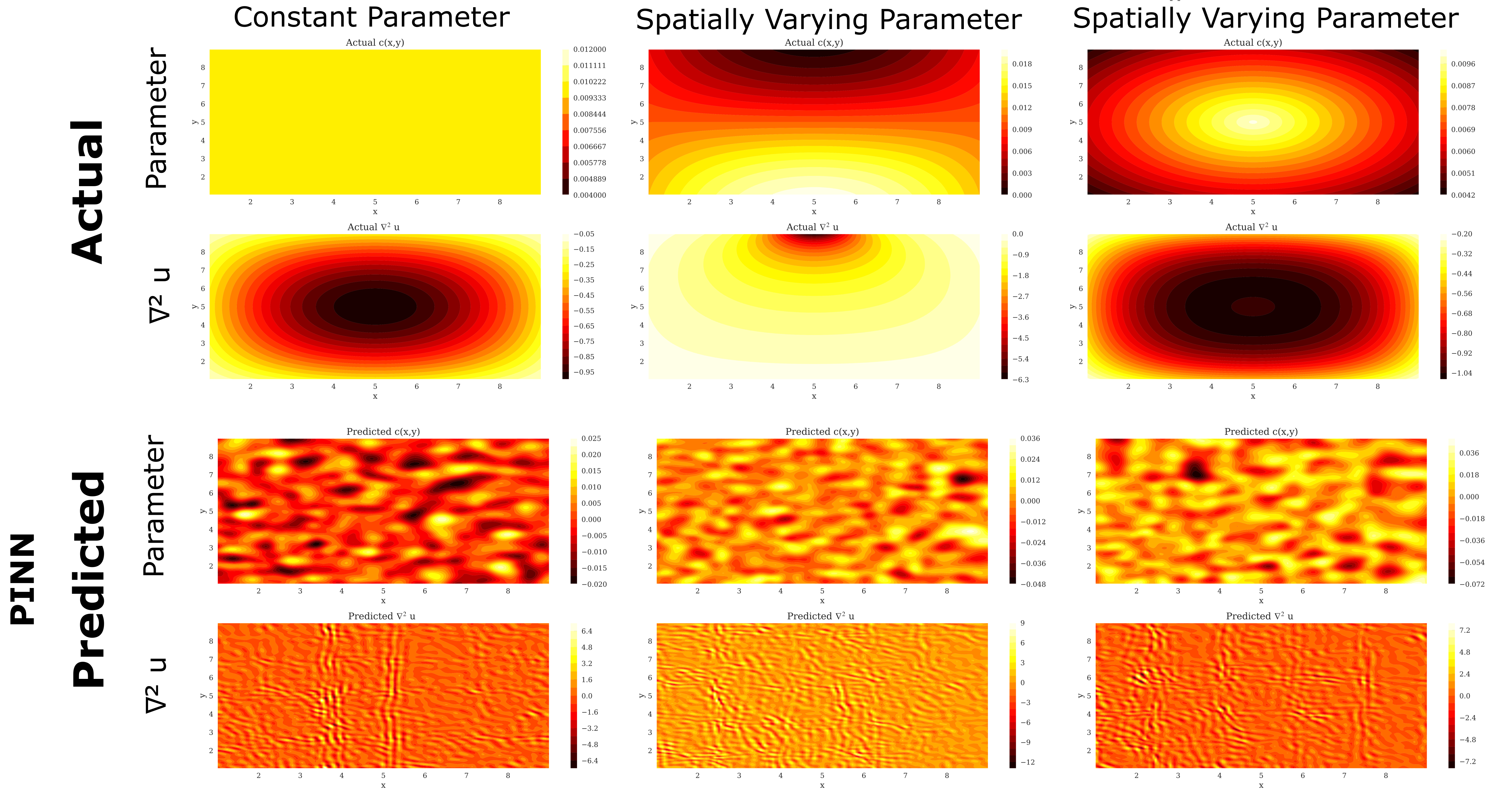}}
\caption{\textit{Estimating constant and spatially varying diffusivity terms and corresponding Laplacians for the heat equation using PINNs}}
\label{icml-historical}
\end{center}
\vskip -0.2in
\end{figure}

\begin{figure}[h]
\vskip 0.2in
\begin{center}
\centerline{\includegraphics[width=\columnwidth]{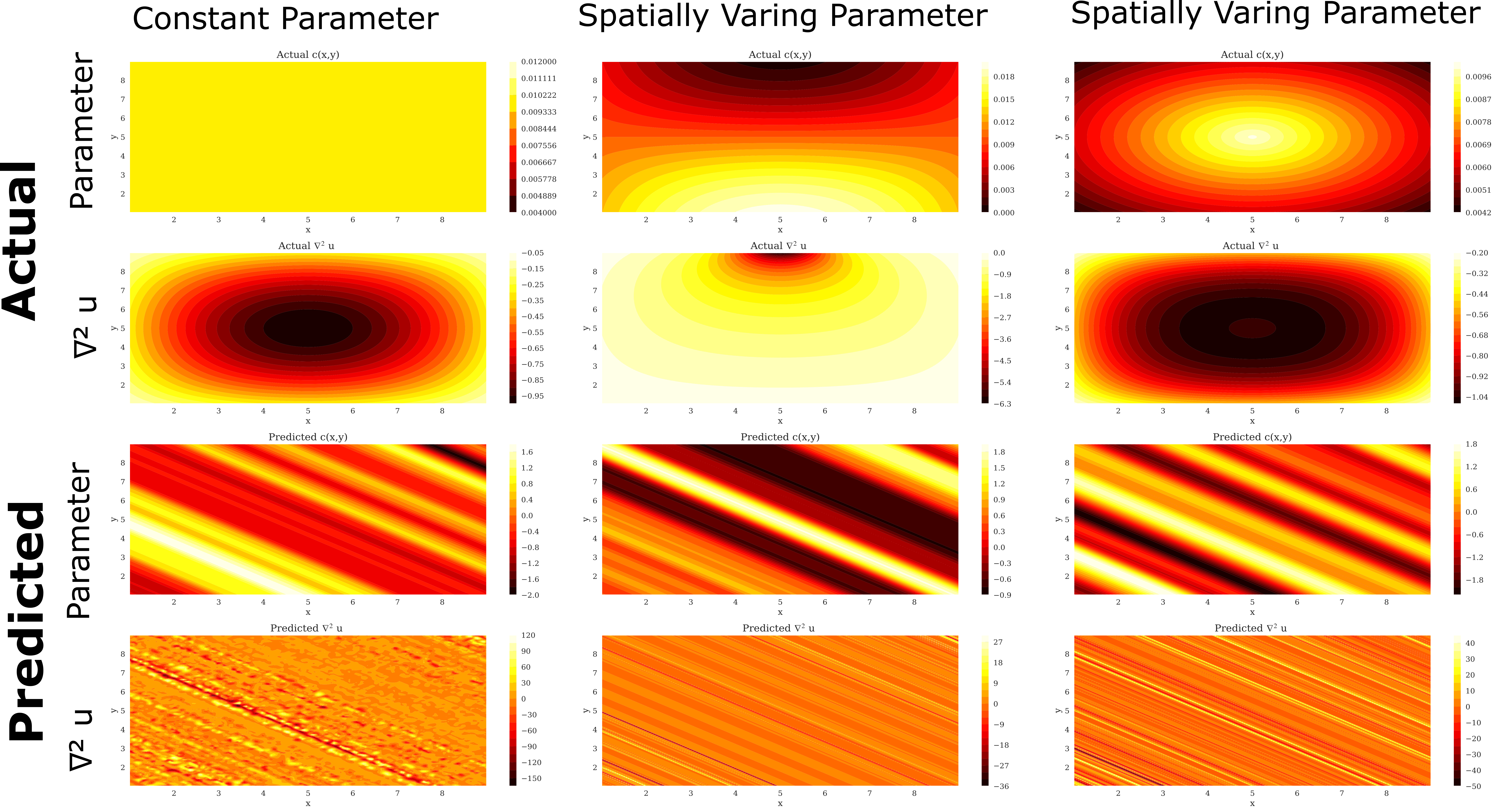}}
\caption{\textit{Estimating constant and spatially varying diffusivity terms and corresponding Laplacians for the heat equation using PirateNets}}
\label{icml-historical}
\end{center}
\vskip -0.2in
\end{figure}

\begin{figure}[h]
\vskip 0.2in
\begin{center}
\centerline{\includegraphics[width=\columnwidth]{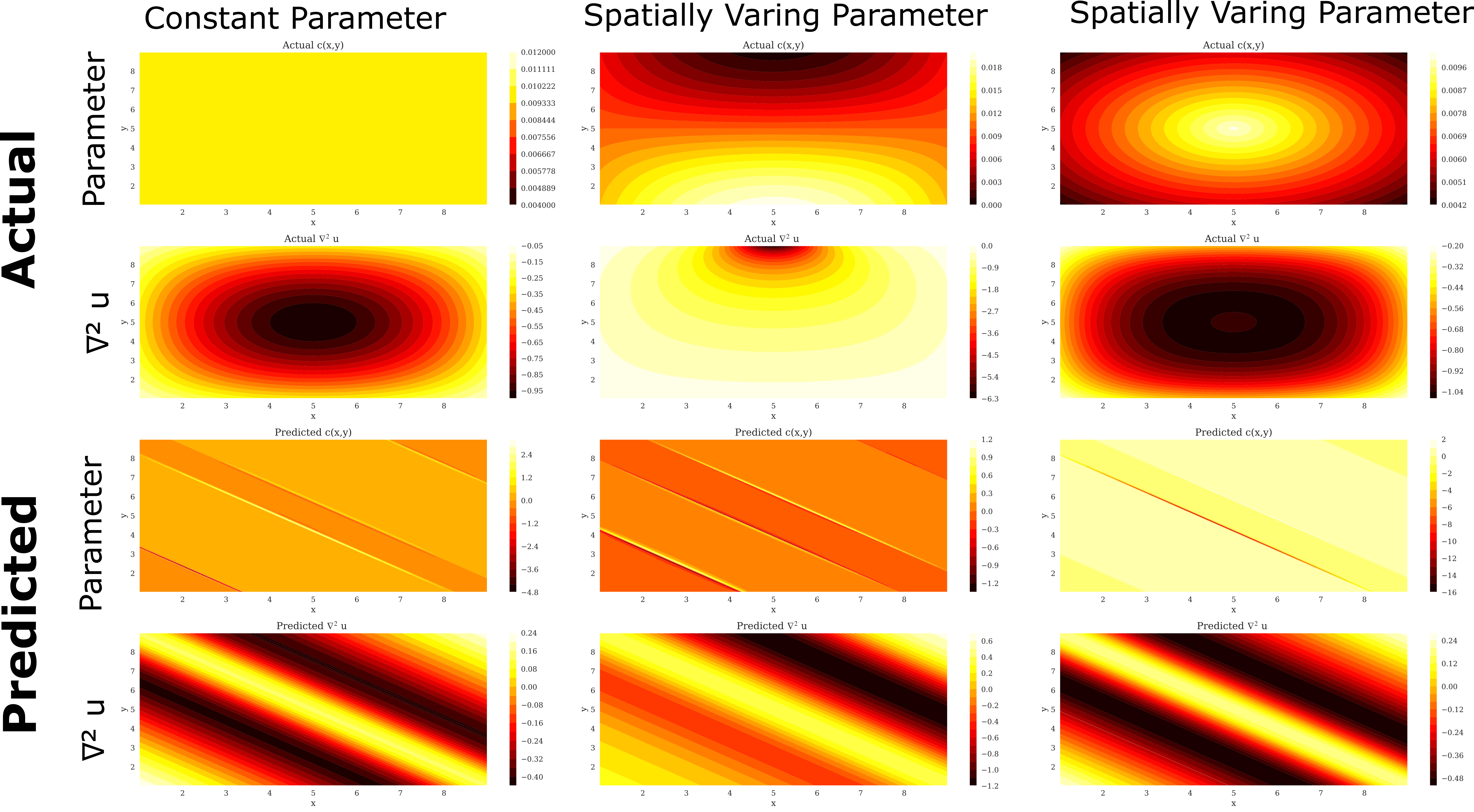}}
\caption{\textit{Estimating constant and spatially varying diffusivity terms and corresponding Laplacians for the heat equation using mollified PirateNets}}
\label{icml-historical}
\end{center}
\vskip -0.2in
\end{figure}

\begin{figure}[h]
\vskip 0.2in
\begin{center}
\centerline{\includegraphics[width=\columnwidth]{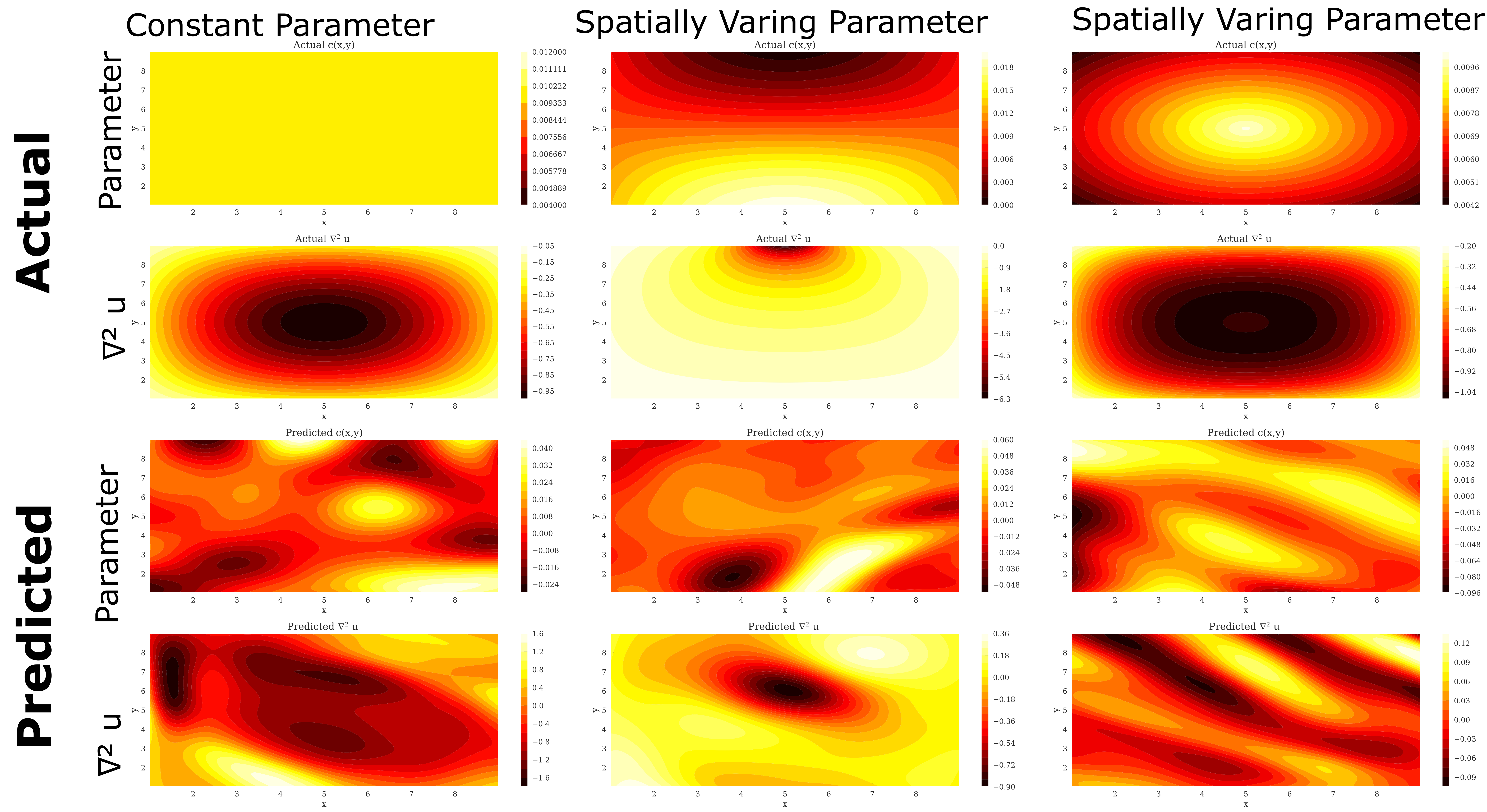}}
\caption{\textit{Estimating constant and spatially varying diffusivity terms and corresponding Laplacians for the heat equation using PINNsFormer}}
\label{icml-historical}
\end{center}
\vskip -0.2in
\end{figure}

\begin{figure}[h]
\vskip 0.2in
\begin{center}
\centerline{\includegraphics[width=\columnwidth]{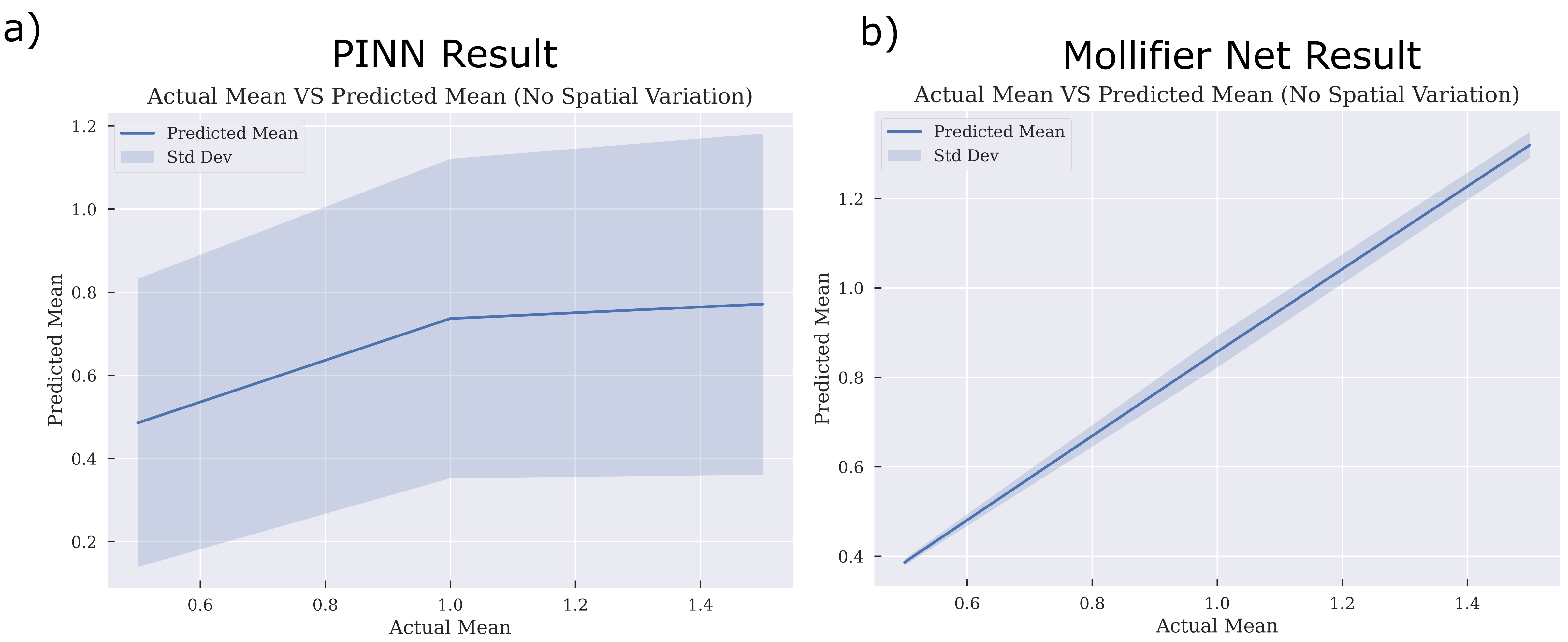}}
\caption{\textit{Predcited vs. Actual mean for the reaction rates in simulations for PINNs and Mollified PINN}}
\label{icml-historical}
\end{center}
\vskip -0.2in
\end{figure}

\begin{figure}[h]
\vskip 0.2in
\begin{center}
\centerline{\includegraphics[width=\columnwidth]{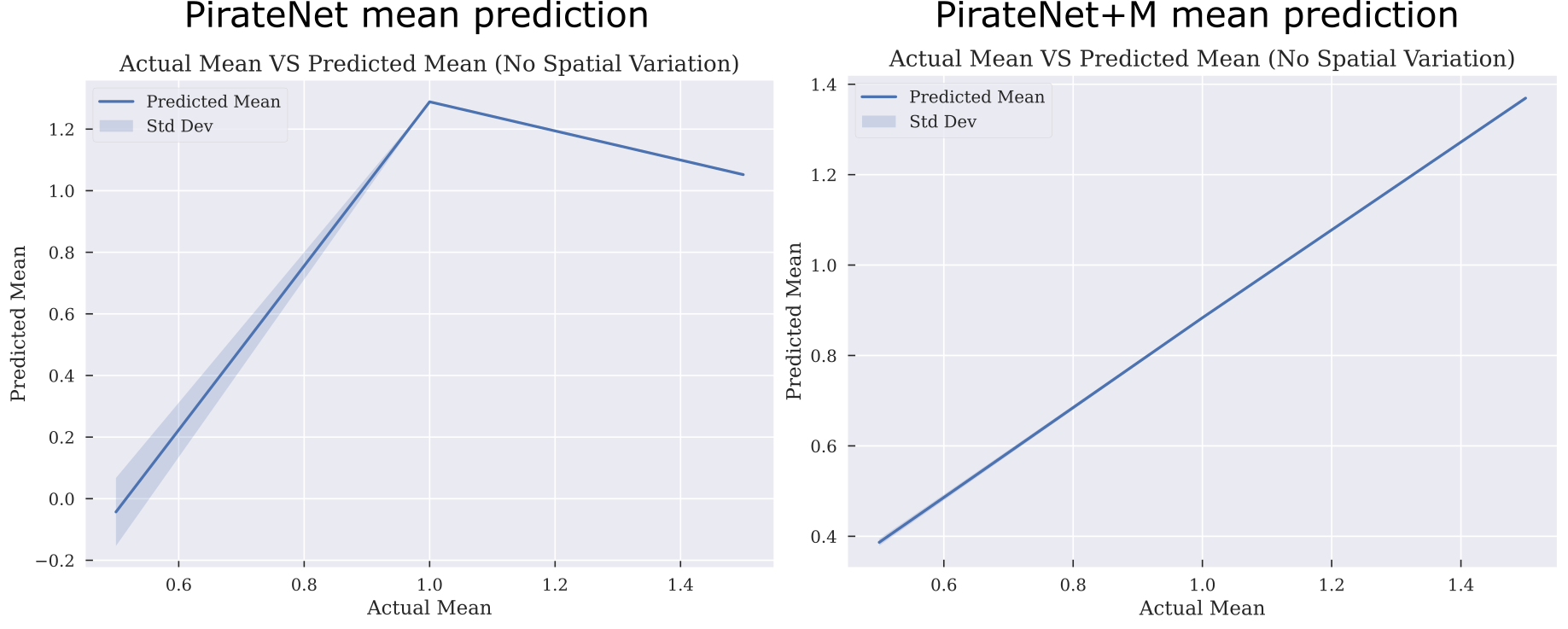}}
\caption{\textit{Predcited vs. Actual mean for the reaction rates in simulations for PirateNet and Mollified PirateNet}}
\label{icml-historical}
\end{center}
\vskip -0.2in
\end{figure}

\begin{figure}[h]
\vskip 0.2in
\begin{center}
\centerline{\includegraphics[width=\columnwidth]{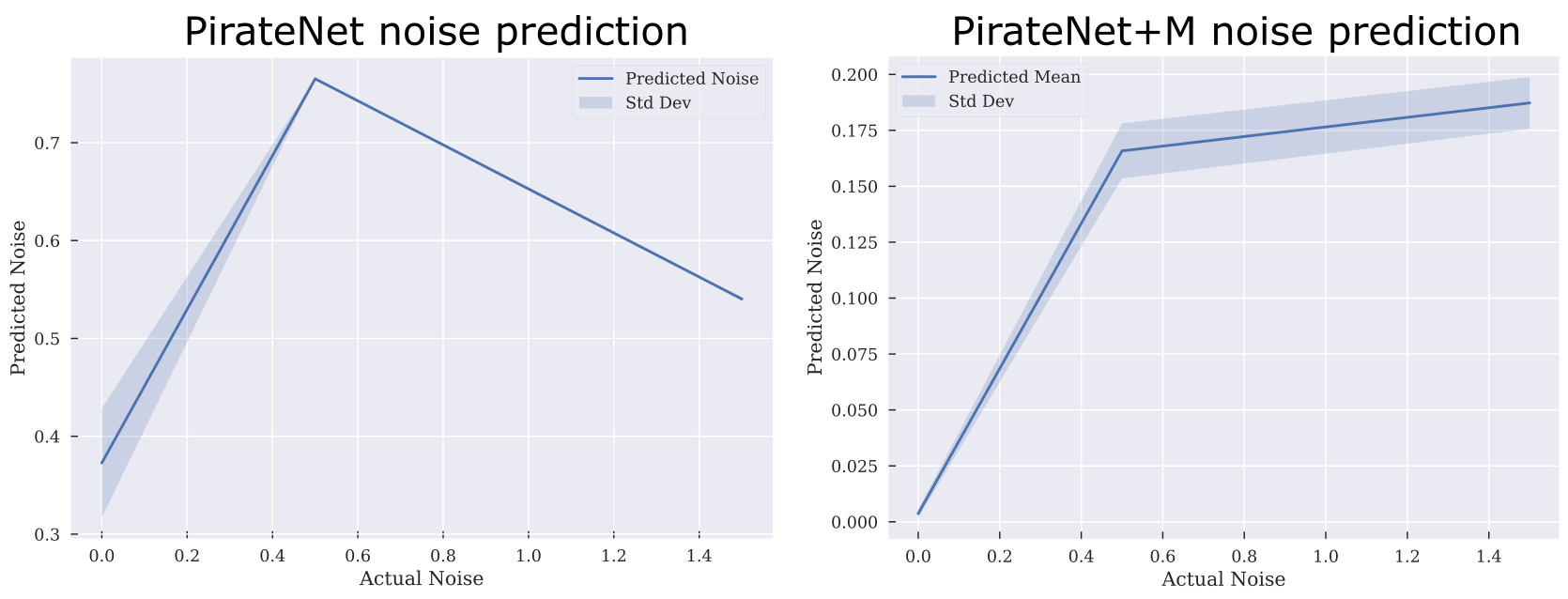}}
\caption{\textit{Predcited vs. Actual noise for the reaction rates in simulations for PirateNet and Mollified PirateNet}}
\label{icml-historical}
\end{center}
\vskip -0.2in
\end{figure}

\begin{figure}[h]
\vskip 0.2in
\begin{center}
\centerline{\includegraphics[width=\columnwidth]{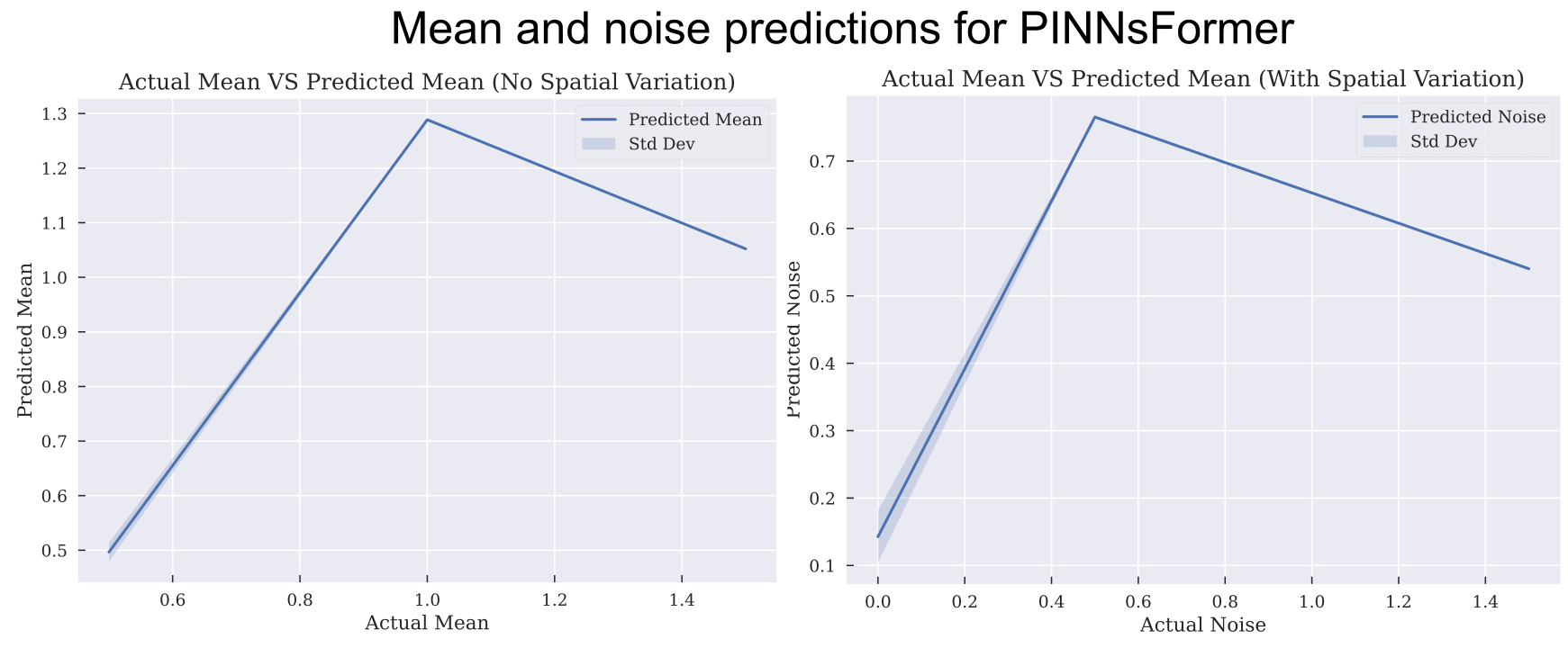}}
\caption{\textit{Predcited vs. Actual mean and noise for the reaction rates in simulations for PINNsFormer}}
\label{icml-historical}
\end{center}
\vskip -0.2in
\end{figure}

\begin{figure}[h]
\vskip 0.2in
\begin{center}
\centerline{\includegraphics[width=\columnwidth]{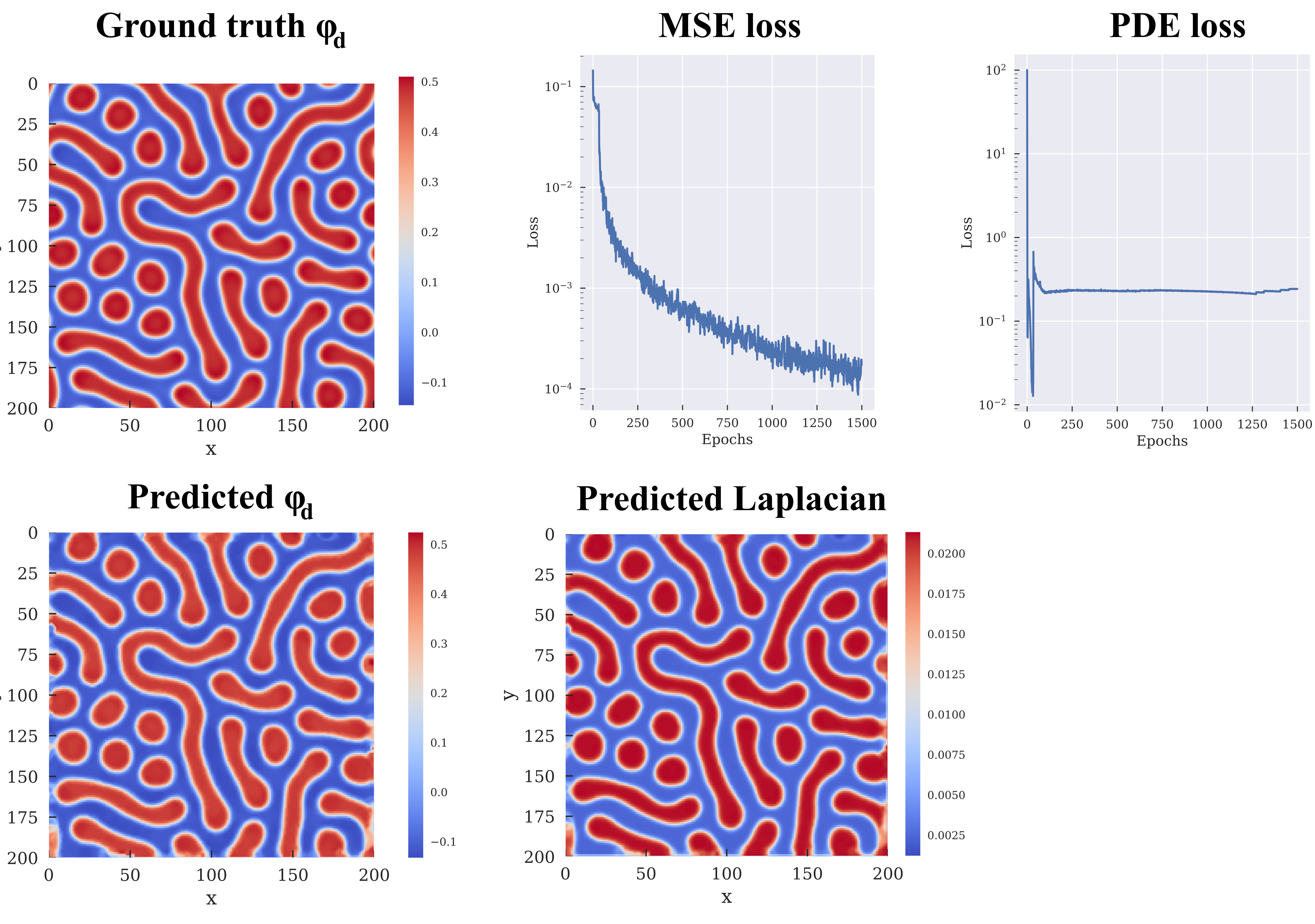}}
\caption{\textit{Mollified PINNs capture the spatial reaction rates for reaction-diffusion equation.}}
\label{icml-historical}
\end{center}
\vskip -0.2in
\end{figure}

\begin{figure}[h]
\vskip 0.2in
\begin{center}
\centerline{\includegraphics[width=\columnwidth]{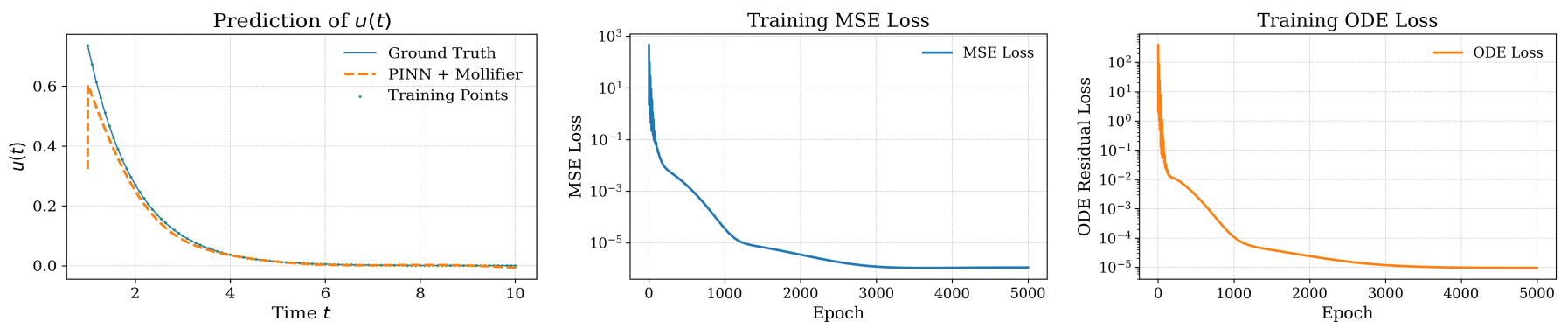}}
\caption{\textit{Using Mollified PINNs for a forward problem.}}
\label{icml-historical}
\end{center}
\vskip -0.2in
\end{figure}

Significant progress has been made in improving PINNs since they first came out\cite{Raissi}, specially showing that gradient flow can be significantly improved by adding residual connections\cite{Wang_gradflow}. Additionally, it is well known that layer normalization helps with gradient smoothening and prevents explosion\cite{xu2019understanding}. Since our aim is to accurately capture gradients and account for this progress made, we used an upgraded PINN, which has the following two features in addition to the base PINN architecture: 

\begin{enumerate}
    
    \item \underline{Residual Connections:} Also known as \textit{skip connections}, residual connections\cite{he2016deep} help mitigate the vanishing gradient problem and facilitate the training of deep neural networks. They can be incorporated into any layer by modifying the standard transformation in a multi-layer perceptron (MLP). Given an input \( x \) to a node in a linear layer \( L \), the standard output \( f(x) \) is defined as:
    \begin{equation}
    f^L(x) = \phi(Wx + b)
    \end{equation}
    
    where \( \phi \), \( W \), and \( b \) represent the activation function, weights, and biases, respectively. To introduce a residual (skip) connection between layer \( L \) and layer \( L+1 \), the modified output \( f^L_{\text{res}}(x) \) is given by:
    \begin{equation}
    f^L_{\text{res}}(x) = \phi(Wx + b) + x
    \end{equation}
    
    This modification allows gradients to propagate more effectively through deeper layers, improving both convergence and model performance.

    \item \underline{Layer Normalization:} Layer normalization\cite{Ba_layernorm} stabilizes and accelerates the training of deep neural networks by normalizing activations across features within each training example, rather than across the batch dimension as in batch normalization. This ensures consistent normalization even for small batch sizes, making it particularly useful in recurrent and transformer-based architectures.
    Given input features \( x \), the output after applying layer normalization is:
    \begin{equation}
    f(x) = \frac{x - \mathbb{E}[x]}{\sqrt{\text{Var}(x) + \epsilon}} \gamma + \beta
    \end{equation}
    
    where \( \mathbb{E}[x] \) and \( \text{Var}(x) \) are the mean and variance computed over the feature dimension of each input, and \( \gamma \) and \( \beta \) are learnable scaling and shifting parameters. This transformation ensures stable feature distributions, improving training convergence and model performance.
    \end{enumerate}

\subsection{Details of the PirateNet implementation}

We adopt the original architectures—with their core components intact—and fine-tune each one by varying their depth specifically for our problems, as detailed in Section B.

\subsection{Details of the PINNsFormer implementation}

We adopt the original architectures—with their core components intact—and fine-tune each one by varying their depth specifically for our problems, as detailed in Section B.

\subsection{Common training choices across models}

We use the following choices for all the models: 
\begin{enumerate}

\item \underline{Fourier Features:} As shown through previous works\cite{wang2021eigenvector}, input features mapped to a higher-dimensional space using \textit{Fourier features} perform better in capturing high frequency signals in the output. Given an input \( x \), we define a set of frequencies \( \omega_i \) within an interval \( I \subset \mathbb{R} \). Specifically, we set \( I = [-3,3] \) and select \( \omega_i \) at equal intervals within this range. The Fourier feature transformation then maps each input \( x \) as follows:
    \begin{equation}
    x \rightarrow [\sin(\omega_1 x), \cos(\omega_1 x), \dots, \sin(\omega_n x), \cos(\omega_n x)]
    \end{equation}
    
    where \( \sin(x) \) and \( \cos(x) \) are applied component-wise to each feature in \( x \).This transformation expands the feature space, allowing the model to capture periodic patterns and high-frequency variations in the data. By incorporating both sine and cosine terms, Fourier features provide a richer oscillatory representation, improving the model’s ability to approximate functions with diverse frequency components. This enhanced representation strengthens the model’s ability to learn and generalize frequency-dependent behaviors.
    
\item \underline{Adam Optimization:} Adam is an adaptive optimization algorithm\cite{kingma2014adam} widely used for training deep learning models. It combines the advantages of \textit{Momentum} and \textit{RMSprop}, adapting the learning rate for each parameter based on both the first moment (mean of gradients) and the second moment (uncentered variance of gradients). This adaptive mechanism allows Adam to efficiently navigate complex optimization landscapes, ensuring faster convergence, especially in problems involving sparse gradients or noisy datasets.  

Adam maintains two moving averages for each parameter \( \theta \):  
\begin{itemize}
    \item \( m_t \): First moment (mean of gradients).
    \item \( v_t \): Second moment (uncentered variance of gradients).
\end{itemize}

At time step \( t \), these moments are updated as follows:  

\begin{equation}
    m_t = \beta_1 m_{t-1} + (1 - \beta_1) g_t
\end{equation}

\begin{equation}
    v_t = \beta_2 v_{t-1} + (1 - \beta_2) g_t^2
\end{equation}

where:  
\begin{itemize}
    \item \( g_t \) is the gradient of the loss function with respect to \( \theta \) at time step \( t \).
    \item \( \beta_1 \) and \( \beta_2 \) are the exponential decay rates for the first and second moments, typically set to \( \beta_1 = 0.9 \) and \( \beta_2 = 0.999 \).
\end{itemize}

To correct the bias toward zero in the initial steps, Adam applies bias correction:  

\begin{equation}
    \hat{m_t} = \frac{m_t}{1 - \beta_1^t}, \quad \hat{v_t} = \frac{v_t}{1 - \beta_2^t}
\end{equation}

Finally, the parameters are updated using:  

\begin{equation}
    \theta_t = \theta_{t-1} - \frac{\alpha}{\sqrt{\hat{v_t}} + \epsilon} \hat{m_t}
\end{equation}

where:  
\begin{itemize}
    \item \( \alpha \) is the learning rate.
    \item \( \epsilon \) is a small constant (typically \( 10^{-8} \)) to prevent division by zero.
\end{itemize}

Adam’s adaptive learning rates make it effective across a wide range of problems, particularly those involving noisy or sparse gradients.

\item \underline{Cosine Learning Rate Decay\cite{Loshchilov_sgdr}:} In deep learning, the learning rate is often initialized at a relatively high value and progressively reduced to refine the model and prevent overshooting minima. A common approach for this adjustment is \textit{cosine annealing}, which schedules the learning rate to follow a cosine curve, ensuring a smooth and gradual decay. Compared to step decay, cosine decay has been shown to improve convergence and generalization.  

The learning rate \( \alpha_t \) at time step \( t \) is defined as:  

\begin{equation}
    \alpha_t =
    \begin{cases} 
        \eta, & \text{if } t < T_0 \\
        \epsilon + \eta \cos \left( \frac{t - T_0}{T - T_0} \right), & \text{if } t \geq T_0
    \end{cases}
\end{equation}

where:  
\begin{itemize}
    \item \( \eta \) is the initial learning rate.
    \item \( \epsilon \) is the minimum learning rate (typically close to zero).
    \item \( T \) is the total number of training iterations or epochs.
    \item \( t \) is the current iteration or epoch.
    \item \( T_0 \) is the point at which the learning rate transitions from constant to decreasing.
\end{itemize}

This decay strategy ensures a smooth reduction in the learning rate, allowing the model to gradually settle into a well-generalized region of the loss landscape. When combined with \textit{Adam optimization}, cosine annealing can lead to faster convergence and improved performance on unseen data.
\end{enumerate}

\subsection{Choice of Mollifying function}

With the constraints of \textit{smoothness, compact support, and non-negativity} outlined in the main text, we define three mollifying functions to explore the behavior of our methodology. We consider a \textit{second-order polynomial}, a \textit{fourth-order polynomial}, and a \textit{sine function} as potential mollification functions, along with the original \textit{exponential mollifier kernel}. These functions are defined as follows:  

Second-order polynomial:
\begin{equation}
    f_2 (x) = -(x^2 - R^2)
\end{equation}

Fourth-order polynomial:
\begin{equation}
    f_4 (x) = -(x^2 - R^2)^2
\end{equation}

Sine function:
\begin{equation}
    f_{\sin} (x) = \sin \left( \frac{\pi R x}{2} + \frac{\pi}{2} \right)
\end{equation}

Original mollification function:
\begin{equation}
    f_{\exp} (x) = e^{\left(\frac{-1}{1 - x^2}\right)}
\end{equation}

where \( x \in B_R (0) \), with \( B_R(0) \) representing the \( L_2 \)-norm ball centered at zero. We refer to \( B_R(0) \) as $U$ in the main text for simplicity.

The kernel size is defined relative to the grid resolution, with values set as follows:  
\begin{itemize}
    \item Langevin equation: \( 0.01 \)
    \item Heat equation: \( 0.005 \)
    \item Reaction-diffusion system: \( 0.05 \)
\end{itemize}

Thus, a kernel size of 7 corresponds to:  
\begin{itemize}
    \item \( R = 0.3 \) for the Langevin equation,
    \item \( R = 0.015 \) for the heat equation,
    \item \( R = 0.15 \) for the reaction-diffusion system.
\end{itemize}

We benchmarked the performance of these kernels using the \textit{Langevin equation with a constant parameter with noise}, with details provided in \textbf{Section B.1}.  

\subsection{Hardware specifications}

Our computing core consists of nodes equipped with four Tesla K80 GPU cards, each containing two GK210GL GPUs, resulting in eight GPUs per node. The GK210GL GPU architecture features 13 Streaming Multiprocessors (SMs), with each SM comprising 192 CUDA cores, yielding a total of 2,496 CUDA cores per GPU. For each training run, we utilize a single node.

\subsection{Weak Form Motivation}
The success of the Finite Element Method (FEM), which employs a weak form approach to solve Partial Differential Equations (PDEs)\cite{brenner2008mathematical}\cite{zienkiewicz2005finite}, has inspired the development of the mollification operation\cite{Friedrichs_weakstrong} presented in this work. Lets begin with function \(v,u, f: U \rightarrow R\). here u is the actual solution to PDE , f is the source term and \(v\) is the weak form approximation of \(u''\).
In the weak form of a PDE, we aim to minimize the residual, which can be expressed as:

\begin{equation}
    \int_{U} v(y) \phi(y) \, dy - \int_{U} f(y) \phi(y) \, dy = 0 \quad \forall \, \phi \in C_c^{\infty}(U)
\end{equation}

We say \( v = u'' \) in the weak sense if,

\begin{equation}
    \int_{U} v(y) \phi(y) = \int_{U} u''(y) \phi(y) \, dy = - \int_{U} u'(y) \phi'(y) \, dy \quad \forall \, \phi \in C_c^{\infty}(U)
\end{equation}

This formulation helps us avoid the direct computation of higher-order derivatives, which would otherwise complicate the process. By approximating the PDE residual with derivatives, we can improve the accuracy of the solution. This leads us to propose an approach in which all derivatives are approximated in the weak form.

To achieve this, we employ a well-established approximation of functions in Sobolev spaces using mollification. This allows us to express the PDE residual in the weak form as:

\begin{equation}
    \left| \int_{U} \eta_{\epsilon}(x - y) v(y) \, dy - \int_{U} \eta_{\epsilon}(x - y) f(y) \, dy \right|
\end{equation}

Here, \( v = D^{\alpha} u \) in the weak sense and \( \eta_{\epsilon}(t) \) is a compact support bounded function in \( (-\epsilon, \epsilon) \). Here $\eta$, also known as the mollifying function, is non-negative, infinitely differentiable and is defined as:
\begin{equation}
    \eta_\epsilon(t)= \eta(\frac{t}{\epsilon})
\end{equation}
\begin{equation}
    \eta(x)= e^\frac{-1}{1-||x||_2^2} \quad \forall ||x||_2 < 1 
    \end{equation}     
\begin{equation}
    \eta(x)= 0 \quad \forall ||x||_2 \geq 1 
\end{equation}

We further relax the above residual by doing the following:

\begin{equation}
    \begin{split}
\Bigl|\!\int_{U}\eta_{\epsilon}(x - y)\,v(y)\,dy 
  - \int_{U}\eta_{\epsilon}(x - y)\,f(y)\,dy\Bigr|
  &\leq \Bigl|\!\int_{U}\eta_{\epsilon}(x - y)\,v(y)\,dy 
    - f(x)\Bigr| \\
  &\quad+\;\Bigl|f(x)
    - \int_{U}\eta_{\epsilon}(x - y)\,f(y)\,dy\Bigr|.
\end{split}
\end{equation}

Note that the residual part of our importance is:

\begin{equation}
    \left| \int_{U} \eta_{\epsilon}(x - y) v(y) - f(x) \right|
\end{equation}

The other term tends to zero as we decrease \( \epsilon \):

\begin{equation}
\begin{split}
\left| f(x) - \int_{U} \eta_{\epsilon}(x - y)\,f(y)\,\mathrm{d}y \right|
&= \left| \int_{U} \eta_{\epsilon}(x - y)\bigl(f(x) - f(y)\bigr)\,\mathrm{d}y \right| \\[0.5ex]
&\le \frac{1}{\epsilon^n}
      \int_{B(x, \epsilon)} 
        \eta\!\Bigl(\tfrac{x - y}{\epsilon}\Bigr)\,
        \bigl|f(x) - f(y)\bigr|\,\mathrm{d}y.
\end{split}
\end{equation}
\begin{equation}
    \frac{1}{\epsilon^n} \int_{B(x, \epsilon)} \eta\left( \frac{x - y}{\epsilon} \right) |f(x) - f(y)| \, dy \leq \frac{C}{\epsilon^n} \int_{B(x, \epsilon)} |f(x) - f(y)|
\end{equation}

where \( C = \sup_{t \in \mathbb{R}^n} \eta(t) \). Using the Lebesgue Differentiation Theorem, we can state that as \( \epsilon \to 0 \):

\begin{equation}
    \frac{1}{\epsilon^n} \int_{B(x, \epsilon)} \left| f(x) - f(y) \right| \, dy \to 0
\end{equation}

Now, we can see that \( v \) is our weak derivative (approximated by a neural network). Thus, the expression becomes:

\begin{equation}
    u_{\text{pred}} = \eta_{\epsilon} \ast \text{NN}_\theta
\end{equation}

Here NN\(_\theta\) is the final output form our Neural network layers. 

As we are working on the grid (G), so we can express the PDE residual with the help of Mollifying operations as:

\begin{equation}
    \frac{\left( \sum_{x \in G} D^{\alpha}_x \eta_{\epsilon} * \text{NN}_\theta(x) - f(x) \right)^2 }{|G|}
\end{equation}
Here $x$ is sampled randomly from a predefined grid of points in the domain. 

\subsection{Convergence Analysis of the Mollifier Layer}

Let $f:[0,1]\to\mathbb{R}$ be continuously differentiable and $L$-Lipschitz. Denote the uniform grid spacing by $h>0$ and the kernel half-width by $\delta>0$. For a nonnegative bump $\rho\in C^{\infty}_c([-1,1])$ with unit mass, define the mollifier
\begin{equation}
(J_{\delta}f)(x)=\int_{-\delta}^{\delta}\rho_{\delta}(r)f(x+r)dr,\qquad
\rho_{\delta}(r)=\delta^{-1}\rho(r/\delta).
\end{equation}
Measured data are $g_j=f(x_j)+\eta_j$ at $x_j=jh$ with $|\eta_j|\le\varepsilon$. Throughout, $C,C_1,C_2,\ldots$ denote positive constants independent of $\delta$, $h$, and $\varepsilon$.

\underline{Mean‑square consistency.\cite{murio1998discrete}} For $f \in L^{2}([0,1])$,
\begin{equation}
\lVert J_{\delta}f - f \rVert_{L^{2}} \to 0 \quad \text{as } \delta \to 0.
\end{equation}

\underline{Uniform consistency\cite{murio1998discrete}.}
\begin{equation}
\lVert J_{\delta}f - f \rVert_{\infty} \le C L \delta, \qquad
\lVert \partial_x J_{\delta}f - f' \rVert_{\infty} \le C L \delta.
\end{equation}

\underline{Discrete mollification\cite{murio1998discrete}.}
\begin{equation}
\lVert J_{\delta}g - f \rVert_{\infty} \le C(\delta + h + \varepsilon).
\end{equation}

\underline{Finite‑difference accuracy\cite{murio1998discrete}.}
\begin{equation}
\bigl|D_0(J_{\delta}g)*j - \partial_x J_{\delta}f(x_j)\bigr| \le C h^{2}, \quad D_0 g_j = \frac{g_{j+1} - g_{j-1}}{2h}.
\end{equation}

\underline{Uniform derivative consistency\cite{murio1998discrete}.}
\begin{equation}
\lVert D_0(J_{\delta}g) - f' \rVert_{\infty} \le C(\delta + h^{2} + \varepsilon).
\end{equation}

Choosing $\delta \approx \sqrt{h}$ and $h^{2} \approx \varepsilon$ minimises the bound, giving $\mathcal{O}(\sqrt{h})$ uniform error while the backward operator norm $\lVert \partial_x J_{\delta} \rVert$ remains $\mathcal{O}(1/\delta)$.

For higher‑order derivatives, differentiating under the integral sign yields
\begin{equation}
\partial_x^{n}(J_{\delta}f) = J_{\delta}^{(n)}f,
\end{equation}
where the superscript acts on the kernel.  The same arguments with a centred stencil of length $2n$ give
\begin{equation}
\lVert \partial_x^{n}(J_{\delta}f) - f^{(n)} \rVert_{\infty} \le C_n(\delta + h + \varepsilon).
\end{equation}

Finally, in the case $n=1$ and absorbing $h^{2}$ into the constants we obtain
\begin{equation}
|\partial_x(J_{\delta}f) - f'|_{\infty} \le C_1\delta + C_2(h + \varepsilon),
\end{equation}
which is the bound quoted in the main text.

\section{Application-wise details}

\subsection{Langevin equation}

\subsubsection{Training Setup for All Models} 

\underline{PINN architechures:} The PINN architecture for predicting \( \hat{u} \) consists of seven layers, each with 1000 nodes and ReLU activation functions. To enhance temporal encoding, we augment the input time \( t \) with Fourier features (details in appendix A). The model is trained using the Adam optimizer with a cosine annealing learning rate schedule. The corresponding mollified PINN maintains the same architecture and training epochs but includes the mollifying layer.  

\underline{PirateNet arch:} PirateNet models the stochastic ordinary differential equation $u_t = u + \lambda(t)$ with an eight-layer fully connected residual network, each layer having 256 hidden units and $\tanh$ activations. The scalar time input $t$ is first lifted into a 256-dimensional Fourier feature embedding (scale factor 2) to capture high-frequency temporal structure. Each layer applies a custom \emph{PI Modified Bottleneck} block: a three-layer MLP whose residual update is mixed via two learnable nonlinear functions $u(\cdot)$ and $v(\cdot)$, and whose skip connection is scaled by a trainable parameter $\alpha$. Finally, two parallel output heads predict the solution $\hat u(t)$ and the auxiliary forcing coefficient $\hat\lambda(t)$, enabling the network to learn both the state evolution and the driving noise in one joint framework.

\underline{PINNsFormer arch:} When applied to the Langevin equation $u_t = u + \lambda(t)$: PINNsformer is configured as a four-layer transformer encoder--decoder with model dimension 64 and four attention heads per layer. The input time $t$ is first encoded using learned sinusoidal embeddings. Each transformer layer applies multi-head self-attention followed by a three-layer feed-forward block activated by WaveAct, a learnable mixture of $\sin(x)$ and $\cos(x)$. Full residual connections and layer normalization are maintained throughout. Two decoder heads then output the state $\hat u(t)$ and the auxiliary estimate $\hat\lambda(t)$, allowing the attention-based model to jointly learn deterministic evolution and stochastic forcing.

All models are trained on a grid sampled from the interval \([0,1]\), using a batch size of 200 for 1000 epochs. The initial learning rate is set to \( 10^{-3} \) with cosine decay scheduling. The Adam optimizer is configured with parameters \( \beta_1 = 0.9 \), \( \beta_2 = 0.999 \).  

\subsubsection{Evaluation on the Langevin Equation}

To test the models, we first infer a constant forcing term for the Langevin equation. As shown in Fig. 5-7, all models successfully predict the constant forcing term. We then extend our analysis to Gaussian white noise predictions, with results summarized in Section 4.1 and Table 2. The figures for all the models are Fig. 8-10. Mollified versions outperform the native nets in all cases.

Thereafter, we also text the models on temporally varying signal along with its noisy version Fig. 13-16. Here too the mollified models outperform the native models.

\subsubsection{Impact of Functional Choices}

The effectiveness of Mollifier-Nets depends on two key parameters:  
\begin{enumerate}
    \item The order of the mollifying function \( \eta \).  
    \item The kernel size, i.e., the integration domain \( U \).  
\end{enumerate}

For the Langevin equation with constant noisy parameter, various configurations accurately estimate the mean \( \Lambda \) (Fig. 11), but they differ in their ability to capture noise variations.  

\underline{Kernel Size Effect:}  
As shown in Fig. 11, for a second-order polynomial mollifier, increasing the kernel size degrades performance on large noise, while a small kernel size fails to capture finer noise variations. An optimal kernel size of 10-15 balances noise capture effectively. This behavior can be attributed to the frequency distribution of the underlying function.  

\underline{Mollifier Order Effect:}  
As shown in Fig. 12, higher-order polynomials perform worse at capturing noise compared to lower-order mollifying functions.  

Conclusively, both polynomial order and kernel size can be optimized. While a low-order polynomial with a moderate kernel size is ideal for extracting Gaussian white noise in the Langevin setting, these choices may vary for different learnable functions.
  
\subsection{Heat equation}

\textsc{Training setup}

\underline{PINN architecture:} The neural network consists of 10 layers, each with 250 nodes and tanh activation functions. Spatial Fourier features are used for positional encoding. Training is performed using the Adam optimizer with a learning rate that progressively decreases following a cosine decay schedule. The mollified version consists of the mollifier layer in addition to the base arch with a original mollifying function.

\underline{PirateNet architecture:}To solve the two-dimensional heat equation, PirateNet uses the same eight residual layers with 256 units each and $\tanh$ nonlinearities. The spatial coordinates $(x,y)$ are first transformed via a 256-dimensional Fourier embedding (scale 2) to enrich spatial frequencies. Inside each PI Modified Bottleneck block, the network mixes the three-layer MLP update with $u(\cdot)$ and $v(\cdot)$ transforms, and applies the learned $\alpha$ scaling to the skip path. The main output head produces the temperature field $\hat u(x,y)$, while an auxiliary head estimates the spatial diffusivity map $\hat\lambda(x,y)$, jointly recovering both solution and physical parameter.

\underline{PINNsFormer architecture:} To handle two-dimensional heat flow, PINNsformer is simplified to a single transformer layer of dimension 32 with one attention head. The spatial coordinates $(x,y)$ are mapped via learned 2D sinusoidal embeddings before entering a WaveAct-activated feed-forward block. Residual connections and normalization persist in this minimal transformer. Its two output heads predict the temperature field $\hat u(x,y)$ and the diffusivity field $\hat\lambda(x,y)$, embedding both solution and parameter estimation within a physics-aware attention framework.

The models are trained on a grid-sampled interval with a batch size of 2500 for 1000 epochs. The initial learning rate is set to \( 10^{-3} \), with Adam optimizer parameters \( \beta_1 = 0.9 \), \( \beta_2 = 0.999 \).

The results are quantitatively summarized in table 2 and the plots are Fig. 18-20.

\subsection{Reaction Diffusion equation}

\subsubsection{Reaction-Diffusion Model for Chromatin Organization}

\title{Reaction-Diffusion Model for Chromatin Organization}
The spatial organization of chromatin within the nucleus plays a fundamental role in gene regulation, influencing cellular function and fate\cite{gilbert2005chromatin}. Chromatin exists in two primary states: heterochromatin, which is compacted and transcriptionally inactive, and euchromatin, which is more open and transcriptionally active. The dynamic interplay between these states is governed by diffusion and biochemical reactions, particularly modifications such as methylation and acetylation\cite{Kant}\cite{vinayak2025polymer}.

\underline{Mathematical Model:}

To model this system, the authors\cite{Kant} consider the nucleus as a mixture of three components: nucleoplasm ($\phi_n$), euchromatin ($\phi_e$), and heterochromatin ($\phi_h$), where their volume fractions satisfy the space-filling constraint:
\begin{equation}
    \phi_h(x,t) + \phi_n(x,t) + \phi_e(x,t) = 1.
\end{equation}

To simplify the system, they define two independent variables:
\begin{itemize}
    \item Nucleoplasm volume fraction ($\phi_n(x,t)$): Represents the proportion of nucleoplasm at a given location.
    \item Chromatin phase difference ($\phi_d(x,t)$): Defined as the difference between heterochromatin and euchromatin fractions:
    \begin{equation}
        \phi_d(x,t) = \phi_h(x,t) - \phi_e(x,t).
    \end{equation}
\end{itemize}
The order parameter $\phi_d(x,t)$ indicates whether a region is euchromatin-rich ($\phi_d < 0$) or heterochromatin-rich ($\phi_d > 0$).

Free Energy Formulation: The free energy density $F(x,t)$ of the system is given by:
\begin{equation}
    F(x,t) = \left[ \phi_e^2 + \phi_h^2 (\phi_h^{\max} - \phi_h)^2 \right] + \frac{\delta^2}{2} |\nabla \phi_n|^2 + \frac{\delta^2}{2} |\nabla \phi_d|^2,
\end{equation}
where:
\begin{itemize}
    \item The first term represents chromatin-chromatin interactions, modeled using a Flory-Huggins-type potential with minima at $\phi_h = 0$ (euchromatin) and $\phi_h = \phi_h^{\max}$ (heterochromatin).
    \item The second term accounts for interfacial energy, penalizing sharp transitions between chromatin phases and promoting smooth boundaries. The parameter $\delta$ controls the interface width.
\end{itemize}

\underline{Diffusion Kinetics of Nucleoplasm:}
The diffusion of nucleoplasm ($\phi_n$) is governed by the gradient of the chemical potential $\mu_n(x,t)$, which drives the system toward a lower-energy configuration. The evolution of nucleoplasm follows the diffusion equation:
\begin{equation}
    \frac{\partial \phi_n}{\partial t} = \nabla^2 \mu_n,
\end{equation}
where $\nabla^2$ is the Laplacian operator, ensuring a conservative dynamics where total nucleoplasm content remains constant unless boundary conditions specify otherwise.

\underline{Reaction-Diffusion Model for Histone Modifications:} Epigenetic modifications such as methylation and acetylation mediate the conversion between euchromatin and heterochromatin. The evolution of the chromatin phase difference ($\phi_d$) is governed by:
\begin{equation}
    \frac{\partial \phi_d}{\partial t} = \nabla^2 \mu_d + 2(\lambda \phi_e - \phi_h),
\end{equation}
where:
\begin{itemize}
    \item $\nabla^2 \mu_d$ represents the diffusive spread of epigenetic modifications.
    \item $\lambda \phi_e - \phi_h$ models methylation-driven conversion from euchromatin to heterochromatin.
\end{itemize}
The chemical potential $\mu_d$ associated with this transition is:
\begin{equation}
    \mu_d = -\phi_e + \phi_h (\phi_h^{\max} - \phi_h)(\phi_h^{\max} - 2\phi_h) - \delta^2 \nabla^2 \phi_d.
\end{equation}
Thus, combining equation 41 and 42 yields a system which is governed by a fourth-order differential equation, indicating complex spatiotemporal dynamics.

\underline{Numerical Solution and Simulations:} To numerically solve the system, we discretize the evolution equations for $\phi_n$ and $\phi_d$, along with their corresponding chemical potentials $\mu_n$ and $\mu_d$. The simulations assume no net exchange of chromatin or nucleoplasm with the surroundings, though boundary conditions can be adjusted to model specific biological scenarios.

\subsubsection{Training setup:} 

\underline{PINN architechure:} The neural network comprises 10 layers, each with 250 nodes using a tanh activation function. We use spatial Fourier features as positional encoding. Training is performed using the Adam optimizer, with a learning rate that progressively decreases during training.

\underline{PirateNet architechure:} \paragraph{PirateNet for the reaction--diffusion system.}
For the reaction--diffusion PDE, PirateNet remains an eight-layer $\tanh$ MLP with 256-unit width. The inputs $(x,y)$ pass through the same 256-dimensional Fourier embedding before entering successive PI Modified Bottleneck blocks, each mixing residual updates via learnable $u$ and $v$ functions and scaling skips by $\alpha$. Two output heads then predict the concentration field $\hat\phi_d(x,y)$ and the reaction-rate field $\hat\lambda(x,y)$, capturing both diffusion dynamics and reaction kinetics in one cohesive residual architecture.

\underline{PINNsFormer architechure:} For reaction-diffusion, PINNsformer again uses one transformer layer of size 32 with a single attention head. Inputs $(x,y)$ receive learned sinusoidal embeddings and pass through a WaveAct feed-forward subblock, all under full residual connections and layer normalization. The model’s main head outputs the concentration field $\hat\phi_d(x,y)$, while an auxiliary head outputs the reaction-rate map $\hat\lambda(x,y)$, integrating physics constraints directly into the attention-based architecture.

The models were trained on a grid sampled on the interval using a batch size of 2500 for 500 epochs. A learning rate of $1 \times 10^{-3}$ with cosine decay scheduling was used. Training employed the Adam optimizer with specific parameters: $\beta_1 = 0.9$, $\beta_2 = 0.999$.

\subsection{Correlation calculations, computing time estimates and memory efficiency estimates}

We assess the accuracy and computational efficiency of our model by measuring the correlation between the actual and predicted parameter values, laplacians and the time/memory usage respectively. These evaluations ensure that our method is not only effective but also feasible for real-world applications where resources such as time and memory are constrained.

\subsubsection{Mean and Noise Calculation}

To evaluate the robustness of our model and assess its sensitivity to statistical fluctuations, we conduct multiple runs on the same problem. Each run produces two key metrics: the predicted mean and the predicted noise.

\underline{Predicted Mean:} The model outputs an estimate of the central tendency of the target variable for each run. This predicted mean represents the model’s best approximation of the expected output given the input features. By averaging the predicted means across multiple runs, we obtain a consolidated estimate of the model’s central tendency.

\underline{Predicted variance:} In addition to the predicted mean, we compute the variability or noise associated with each prediction. By averaging these predicted noise values, we gain insights into the uncertainty of the model's predictions.

\underline{Correlation Analysis:}
Once the actual and predicted mean, variance and the laplacians are computed, the next step is to assess the correlation between the actual and predicted values. This ensures that the model effectively captures underlying trends and relationships in the data.

To quantify this relationship, we compute the \textbf{correlation coefficient}, defined as:

\begin{equation}
    r = \frac{\sum\limits_{i=1}^{n} (X_i - \bar{X})(Y_i - \bar{Y})}
    {\sqrt{\sum\limits_{i=1}^{n} (X_i - \bar{X})^2 \sum\limits_{i=1}^{n} (Y_i - \bar{Y})^2}}
\end{equation}

where:
\begin{itemize}
    \item \(X_i\) and \(Y_i\) represent the actual and predicted values, respectively.
    \item \( \bar{X} \) and \( \bar{Y} \) denote the mean of the actual and predicted values.
\end{itemize}

This correlation coefficient provides a quantitative measure of the strength and direction of the relationship between actual and predicted values.

\subsubsection{Time Evaluation}

To measure the execution time of our model, we use Python’s built-in \texttt{time} module, which provides a simple way to capture the start and end times of a computation. By recording the time before and after running the model, we can compute the total time taken for a single run. This approach allows us to evaluate the speed of our model under various conditions and for different input sizes. 

Additionally, execution time measurements across different experiments enable the identification of performance bottlenecks and an assessment of whether the model’s time complexity scales appropriately with increasing data or model size.

\subsubsection{Memory Evaluation}

To measure memory consumption, we utilize PyTorch’s built-in utilities for tracking memory usage during training and inference. Specifically, when using GPUs, we employ the functions \texttt{torch.cuda.memory\_allocated()} and \texttt{torch.cuda.memory\_reserved()} to monitor memory allocation. 

By tracking both GPU and system memory consumption, we can determine whether the model is memory-efficient and ensure that it operates within available system resources without causing memory overflow or excessive usage.

\subsection{Forward toy problem}

While our focus is on inverse problems, as demonstrated extensively throughout this work, we briefly illustrate the applicability of mollifier layers to a forward problem. The forward problem is governed by the linear ordinary differential equation $\frac{d}{dx} u(x) + \lambda u(x) = 0$, with $\lambda = 1.0$. The solution $u(x)$ is approximated using a fully connected neural network with hidden layer of 64 neurons each and Tanh activations. To compute spatial derivatives in a stable and differentiable manner, we employ mollifier-based convolutions. The loss function consists of two terms: an ODE loss that enforces the differential equation, and a data loss that aligns the mollified prediction with the sampled data. We use a mollifier kernel of size 11 with a smoothing radius $\varepsilon = 5 \Delta x$, where the spatial discretization $\Delta x = 0.05$. The model is trained using the Adam optimizer with cosine learning rate decay over 5000 epochs. This example illustrates how mollifier layers can be used in the forward problem. The predictions as well as the learning curves are provided in Fig. 26.

\section{Super-resolution image processing}

Since the super-resolution images are binary, they only indicate the presence or absence of a signal at a given locus. The density of these activated loci reflects the chromatin density at a given position, where a higher density corresponds to more heterochromatin. To establish a correspondence between the imaging data and the reaction-diffusion system, it is necessary to determine a density equivalent to the obtained super-resolution imaging data. In this work, we define a new square grid to better identify the distribution of chromatin within the nucleus. To compute the density of chromatin at a given point on this grid, we use the following methodology: 

Watson Kernel interpolation\cite{watson1964smooth} is employed to perform spatial interpolation of data points in super-resolution images. This method utilizes a Gaussian kernel to compute a weighted average estimate of nearby points, ensuring smooth interpolation over the image grid. The Gaussian kernel is defined as:

\begin{equation}
    K(d) = \frac{\exp\left(-\frac{d^2}{2\sigma^2}\right)}{\sqrt{2\pi\sigma^2}}
\end{equation}

where \( d \) represents the Euclidean distance between points, and \( \sigma \) is the kernel's bandwidth parameter, which controls the kernel width. The kernel assigns higher weights to closer points, effectively smoothing the interpolation based on proximal density.

Then the Nadaraya-Watson estimator \cite{nadaraya1964estimating}\cite{watson1964smooth} is used to compute the weighted average of data values, where the weights are determined by the above Gaussian kernel. Specifically, for a query point \( (x_{\text{new}}, y_{\text{new}}) \), the estimated value \( z_{\text{new}} \) is calculated as:

\begin{equation}
    z_{\text{new}} = \frac{\sum\limits_{i=1}^{n} w_i z_i}{\sum\limits_{i=1}^{n} w_i}
\end{equation}

where \( w_i \) is the weight assigned to each data point based on its distance to \( (x_{\text{new}}, y_{\text{new}}) \), and \( z_i \) is the corresponding value at each data point.

For computational efficiency, \textit{batch processing} is used to perform interpolation in chunks. This method divides the target grid into smaller batches and computes the weighted average for each batch, ensuring that the interpolation process scales efficiently with large datasets. The interpolation is implemented using the following steps:

\begin{enumerate}
    \item Compute the distances between target points and known data points.
    \item Apply the Gaussian kernel to compute weights.
    \item Calculate the weighted sum of the data values.
    \item Normalize the weighted sum by the total weights to obtain the interpolated values.
\end{enumerate}

After applying Watson Kernel interpolation, the interpolated values are used to update image features such as \( \phi_h \), \( \phi_e \), and \( \phi_n \). These terms are computed as:

\begin{align}
    \phi_h &= \left(\frac{z_{\text{mesh}} \times 6}{7}\right)^{0.5}, \\
    \phi_n &= \left(\frac{z_{\text{mesh}} - 0.7}{0.7}\right)^2 \times \frac{6}{7 \times 0.7} + 0.2, \\
    \phi_e &= 1 - \phi_h - \phi_n.
\end{align}

These processed values are then used for further analysis, providing refined estimates of the underlying super-resolution imaging features.

\end{document}